\documentclass{article}

\usepackage[final,nonatbib]{neurips_data_2021}

\usepackage[utf8]{inputenc} 
\usepackage[T1]{fontenc}    
\usepackage{hyperref}       
\usepackage{url}            
\usepackage{booktabs}       
\usepackage{amsfonts}       
\usepackage{nicefrac}       
\usepackage{microtype}      
\usepackage{float}
\usepackage{makecell}
\usepackage[table]{xcolor}
\usepackage{subcaption}
\usepackage[numbers]{natbib}
\usepackage{graphicx}
\usepackage{wrapfig}
\usepackage{adjustbox}
\usepackage{relsize}
\usepackage{enumitem}

\usepackage{pifont} 
\definecolor{mygreen}{RGB}{34,139,34}
\newcommand{\y}{\textcolor{mygreen}{\ding{51}}}

\newcommand{\n}{\textcolor{red}{\ding{55}}}

\usepackage{ifthen}
\newboolean{include-notes}
\setboolean{include-notes}{true}

\newcommand{\TODO}[1]{\ifthenelse{\boolean{include-notes}}
 {{\color{red} TODO: #1}}{}}
\newcommand{\Rohin}[1]{\ifthenelse{\boolean{include-notes}}
 {{\color{mygreen} RS: #1}}{}}
\newcommand{\Cody}[1]{\ifthenelse{\boolean{include-notes}}
 {{\color{magenta} CW: #1}}{}}
\newcommand{\Cynthia}[1]{\ifthenelse{\boolean{include-notes}}
 {{\color{blue} CC: #1}}{}}
\newcommand{\Sam}[1]{\ifthenelse{\boolean{include-notes}}
 {{\color{cyan} ST: #1}}{}}

\newcommand{\prg}[1]{\noindent\textbf{#1}}

\newcommand{\expect}[2]{\mathop{\mathbb{E}}_{#1}\left[{#2}\right]}

\newcommand{\loss}[1]{\mathcal{L}_{\text{#1}}}
\newcommand{\given}{\mid}

\usepackage{amsmath,amsfonts,bm}

\def\eqref#1{equation~\ref{#1}}

\def\1{\bm{1}}

\newcommand{\mpm}{\mathsmaller{\pm}}

\DeclareMathAlphabet{\mathsfit}{\encodingdefault}{\sfdefault}{m}{sl}
\SetMathAlphabet{\mathsfit}{bold}{\encodingdefault}{\sfdefault}{bx}{n}

\usepackage[capitalize]{cleveref}

\newcommand{\dmc}{{DMC}}

\title{An Empirical Investigation of \\ Representation Learning for Imitation}

\author{%
  Xin Chen\thanks{Equal contribution, corresponding authors} \\
  The University of Hong Kong \\
  \texttt{cyn0531@connect.hku.hk} \\
  \And 
 Sam Toyer\footnotemark[1] \\
  UC Berkeley \\ 
  \texttt{sdt@berkeley.edu} \\
 \And 
Cody Wild\footnotemark[1] \\
UC Berkeley \\ 
\texttt{codywild@berkeley.edu} \\
 \And 
Scott Emmons \\
UC Berkeley \\ 
 \And 
Ian Fischer \\
Google Research \\ 
 \And 
Kuang-Huei Lee \\
Google Research \\ 
 \And 
Neel Alex \\
UC Berkeley \\ 
 \And 
Steven Wang \\
UC Berkeley \\ 
\And 
Ping Luo \\
The University of Hong Kong \\ 
\And 
Stuart Russell \\
UC Berkeley \\ 
\And 
Pieter Abbeel \\
UC Berkeley \\ 
\And 
Rohin Shah \\
UC Berkeley \\ 
}

\begin{document}

\maketitle

\begin{abstract}
Imitation learning often needs a large demonstration set in order to handle the full range of situations that an agent might find itself in during deployment. However, collecting expert demonstrations can be expensive. Recent work in vision, reinforcement learning, and NLP has shown that auxiliary representation learning objectives can reduce the need for large amounts of expensive, task-specific data. Our Empirical Investigation of Representation Learning for Imitation (EIRLI) investigates whether similar benefits apply to imitation learning. We propose a modular framework for constructing representation learning algorithms, then use our framework to evaluate the utility of representation learning for imitation across several environment suites. In the settings we evaluate, we find that existing algorithms for image-based representation learning provide limited value relative to a well-tuned baseline with image augmentations. To explain this result, we investigate differences between imitation learning and other settings where representation learning \textit{has} provided significant benefit, such as image classification. Finally, we release a well-documented codebase which both replicates our findings and provides a modular framework for creating new representation learning algorithms out of reusable components.
\end{abstract}

\section{Introduction}


Much recent work has focused on how AI systems can learn what to do from human feedback~\cite{jeon2020reward}. The most popular approach---and the focus of this paper---is \textit{imitation learning} (IL), in which an agent learns to complete a task by mimicking demonstrations of a human.


As demonstrations can be costly to collect, we would like to learn representations that lead to better imitation performance given limited data. Many existing representation learning (RepL) methods in Computer Vision and Reinforcement Learning do exactly this, by extracting effective visual~\cite{chen2020simple} or temporal~\cite{lee2020predictive} information from inputs. A natural hypothesis is that RepL would also add value for IL.


We test this hypothesis by investigating the impact of common RepL algorithms on Behavioral Cloning (BC) and Generative Adversarial Imitation Learning (GAIL). We survey a wide variety of RepL methods, and construct a modular framework in which each design decision can be varied independently. As previous work has found that image augmentation alone can outperform more complex representation learning techniques~\cite{laskin2020reinforcement, kostrikov2020image}, we make sure to compare against baselines that use augmentation. To ensure generalizability of our results, we evaluate on ten tasks selected across three benchmarks, including MAGICAL~\cite{toyer2020magical}, Procgen~\cite{cobbe2020leveraging} and the DeepMind Control Suite (DMC)~\cite{tassa2018deepmind}. 

We find that, on average, RepL methods do significantly outperform vanilla BC, but this benefit can be obtained simply by applying well-tuned image augmentations during BC training. To understand the discrepancy between this result and the success of RepL in computer vision and reinforcement learning, we apply clustering algorithms and attribution methods to qualitatively investigate the learned representations and policies, surfacing a number of intriguing hypotheses for investigation in future work.

This paper is, to the best of our knowledge, the first to provide a systematic empirical analysis of different representation learning methods for imitation learning in image-based environments. Concretely, our Empirical Investigation of Representation Learning for Imitation (EIRLI) makes the following contributions:

\begin{enumerate}[leftmargin=15pt]
\item We identify meaningful axes of variation in representation learning algorithm design, allowing us to construct a modular framework to conceptually analyze these designs.
\item We use this framework to build a well documented, modular, and extensible code base, which we release at \href{https://github.com/HumanCompatibleAI/il-representations/}{\texttt{github.com/HumanCompatibleAI/eirli}}. 
\item We conduct an extensive comparison of popular RepL methods in the imitation learning setting, and show that RepL has limited impact on task performance relative to ordinary image augmentations.
By analysing our learned representations and policies, we identify several promising directions for future work at the intersection of representation learning and decision-making.
\end{enumerate}

\section{Design decisions in representation learning}
\label{sec:repl-analysis}

\begin{figure}
    \centering 
    \includegraphics[trim={0.2cm 3.6cm 0.2cm 0.2cm},width=\textwidth]{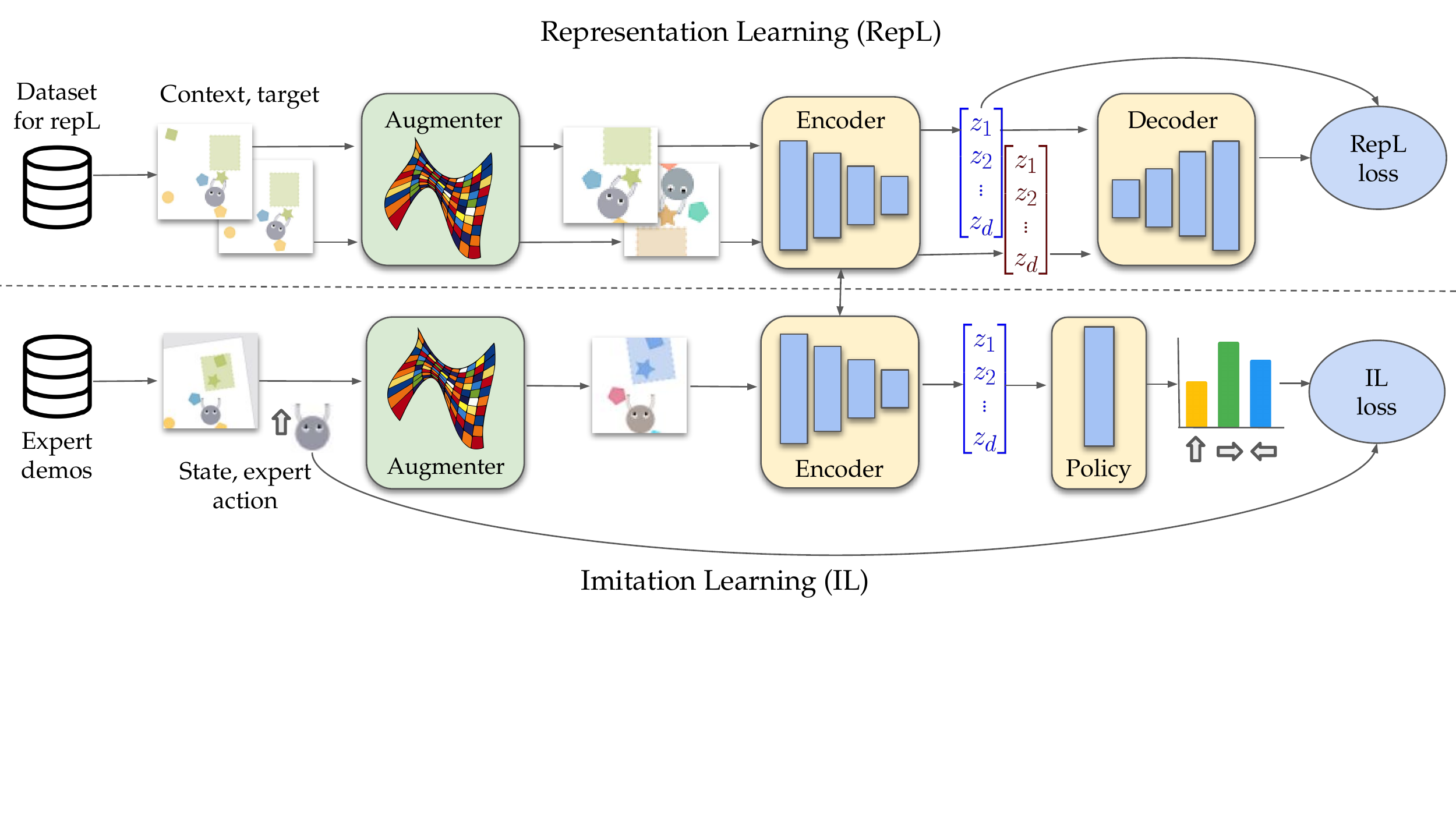}
     \caption{A framework for the use of representation learning (RepL) in imitation learning. In the pretraining setting, we first train the encoder with RepL, then finetune end-to-end with IL. In the joint training setting, the RepL objective is used as an auxiliary loss throughout IL training.}
\label{fig:pipeline_diagram}
\end{figure}

To apply representation learning (RepL) effectively, it is important to understand the relative impact of different RepL algorithm design choices on downstream task performance. We argue that for many common RepL algorithms, these design choices can be broken down along a common set of axes, which we show in \cref{table:existing-algorithms-vision} and \cref{table:existing-algorithms-rl-summary}. In this section, we elaborate on our conceptual breakdown both as a literature review and as an implementation walkthrough of our RepL framework.

We summarize existing RepL for image classification algorithms in Table~\ref{table:existing-algorithms-vision} and a selection of RepL for reinforcement learning algorithms in Table~\ref{table:existing-algorithms-rl-summary}. The full version of the table deconstructing current RepL methods in reinforcement learning can be found in the appendix in Table~\ref{table:existing-algorithms-rl}.
\cite{kostrikov2020image}

\begin{table*}[t]
\caption{Design choices made in representation learning for image recognition. ``Augmentation'', ``Momentum'', and ``Projection'' show whether image augmentation, target encoder momentum, and projection heads were used, respectively. ``Pre/Joint'' shows whether RepL is used as a pretraining step, or is jointly learned with the downstream task (typically as an auxiliary loss).}
\label{table:existing-algorithms-vision}
\begin{center}
\begin{tabular}{@{}lccccccc@{}}
\toprule
{\bf Algorithm}                             & {\bf Task}    & {\bf Augmentation} & {\bf Momentum} & {\bf Projection} & {\bf Pre/Joint} \\
\midrule
VAE~\cite{kingma2013auto}           & Reconstruction       & \n  & \n  & \n  & Pre  \\
AugMix~\cite{hendrycks2019augmix}   & Consistency          & \y  & \n  & \n  & Joint  \\
FixMatch~\cite{sohn2020fixmatch}    & Consistency          & \y  & \n  & \n  & Joint  \\
CPC~\cite{oord2018representation}   & Contrastive          & \n  & \n  & \y  & Pre  \\
MoCo~\cite{he2020momentum}                 & Contrastive   & \y  & \y  & \n  & Pre  \\
SimCLR~\cite{chen2020simple}               & Contrastive   & \y  & \n  & \y  & Pre  \\
SimCLRv2~\cite{chen2020big}                & Contrastive   & \y  & \n  & \y  & Pre  \\
BYOL~\cite{grill2020bootstrap}             & Bootstrap     & \y  & \y  & \y   & Pre  \\
\bottomrule
\end{tabular}
\vspace{-5mm}
\end{center}
\end{table*}

\setlength{\tabcolsep}{2pt} 
\begin{table}[H]
\caption{Design choices made in a selection of representation learning algorithms for reinforcement learning (full table in the appendix). Act, Aug, Mom, Proj and Comp respectively show whether action conditioning, augmentation, momentum, projection heads, and compression were used. P/J determines whether the representation learning is an initial (P)retraining step, or is (J)ointly learned alongside reinforcement learning. R/C in the Task column refer to Reconstruction/Contrastive. Note that different papers may use different sets of augmentations.}
\label{table:existing-algorithms-rl-summary}
\begin{center}
\begin{tabular}{@{}lcccccccccc@{}}
\toprule
{\bf Algorithm} & {\bf Task}    & {\bf RL alg.}  & {\bf Context} & {\bf Target}          & {\bf Act} & {\bf Aug} & {\bf Mom} & {\bf Proj} & {\bf Comp} & {\bf P/J} \\
\midrule
World models~\cite{ha2018world}   & R   & CMA-ES    & $o_t$         & $o_t, o_{t+1}$        & \y  & \n  & \n  & \n  & \n  & P \\
PlaNet~\cite{hafner2019learning}          & R   & MPC + CEM & $o_{1:t}$     & $o_{t+1:T}, r_{t+1:T}$& \y  & \n  & \n  & \n  & \n  & J \\
CURL~\cite{laskin2020curl}            & C      & SAC       & $o_t$         & $o_t$                 & \n  & \y  & \n  & \n  & \n  & J \\
PI-SAC~\cite{lee2020predictive}   & C      & SAC       & $o_t$         & $o_{t+k}, r_{t+k}$    & \y  & \y  & \y  & \y  & \y  & J \\
ATC~\cite{stooke2020decoupling}      & C      & SAC, PPO         & $o_t$         & $o_{t+k}$             & \n  & \y  & \y  & \y  & \n  & P \\
\bottomrule
\end{tabular}
\end{center}
\end{table}

\subsection{Target selection}
Most RepL methods can be thought of as proxy tasks in which a dataset of $(x, y)$ pairs is provided and the network must model some aspects of the relationship between $x$ and $y$. Since the learning signal derives from the relationship between $x$ and $y$, the choice of $x$ and $y$ thus has a significant impact on exactly what information is modeled. We refer to the inputs for which representations $z$ are computed as the ``context'' $x$, and the inputs with which they are related are the ``targets'' $y$.  Often, the target is a (possibly transformed) version of a context.

In image classification, learned representations must capture the label-relevant information in a single input image. It is assumed that most images used for representation learning will not have labels or other task-relevant metadata. Thus, the context and target are typically both set to the original image, after which they may be augmented in different ways. For example, in a Variational Autoencoder (VAE) \cite{kingma2013auto}, an input image (context) is encoded into a vector representation and then decoded back into pixels, which is then compared against the same input image (now interpreted as a target).

Once we move to sequential decision-making, the observations have a sequential structure, and there is a notion of actions and a reward function. These can all be leveraged in the construction of the contexts and targets. For example, a \emph{Temporal VAE} is identical to a regular VAE, except that for a context observation $o_t$, we set the target to be a future observation $o_{t+k}$. Now, the input observation $o_t$ (context) is encoded into a vector representation and then decoded back into pixels, which is then compared against the \emph{future} observation $o_{t+k}$ (target). By using a temporal target, we now incentivize representations that contain \emph{predictive} information~\cite{schmidhuber1991reinforcement}.
In reinforcement learning, another option is to add the reward $r_t$ to the target to encourage learning representations that are useful for planning.


\subsection{Loss type}
We divide modern methods for representation learning into four categories:

\prg{Reconstruction.} Here, the goal is to reconstruct the target $y$ from the representation $z$. Both the VAE and temporal VAE in the previous section use a reconstructive loss, in which a \emph{decoded image} $d_{\phi}(z)$ is compared against the target $y$, and that reconstruction loss is combined with a regularization term.

\prg{Contrast.} Contrastive methods take a series of context--target pairs $(x_1, y_1), (x_2, y_2), \ldots, (x_K, y_K)$ and use the same network to encode both the context and target into latent representations $z_i \sim e(z \given x_i)$ and $z_i' \sim e(z \given y_i)$.
A contrastive loss then incentivizes $z_i$ and $z_i'$ to be similar to each other, but different from $z_j$ and $z_j'$ for all other pairs $j \neq i$.
Typically, the contrastive loss function is chosen to maximize the mutual information $I(z ; y)$, such as with the InfoNCE loss function~\cite{poole2019variational}:
\[ \loss{InfoNCE} = \mathbb{E}\left[\log \frac{e^{f\left(x_{i}, y_{i}\right)}}{\frac{1}{K} \sum_{j=1}^{K} e^{f\left(x_{i}, y_{j}\right)}}\right] \]
$f$ could, for instance, be a bilinear function $f(x_i, y_i) = z_i^T W z_i'$, where $z_i \sim e(z \given x_i)$, $z_i' \sim e(z \given y_i)$, and $W \in \mathbb R^{n \times n}$ is a learned parameter matrix.

\prg{Bootstrapping.} This is a simplified variant of contrastive learning. Given a related context $x$ and target $y$, a bootstrapping method predicts a moving-average-encoded target from the encoded context.
Bootstrapping does not need a large dataset of negatives to prevent the representation from collapsing to a single point; instead, it prevents collapse by stopping gradients from propagating through the target encoder.

\prg{Consistency.} These methods, such as AugMix~\cite{hendrycks2019augmix} and FixMatch~\cite{sohn2020fixmatch}, include auxiliary loss terms that encourage the model to produce similar distributions over $y$ for different transformations of the same input image.

\prg{Compression.} A representation $z \sim e_\theta(\cdot \given x)$ should contain enough information about the input $x$ to solve downstream tasks. Ideally, $e_\theta$ should also extract only the \textit{minimum} amount of information about $x$ that is necessary to perform well.
We refer to this as \textit{compression}. As a form of explicit compression, we implement the \emph{conditional entropy bottleneck} (CEB)~\cite{fischer2020conditional}, which approximately minimizes $I(X ; Z \given Y)$.

\subsection{Augmentation}
In many algorithm designs, one or both of the context frame and target frame undergo augmentation before being processed by the encoder and decoder networks. In some algorithms, like SimCLR, this augmentation is the main source of noise causing transformed representations of the same input to not be purely identical. In other algorithms, it simply helps promote generalization by sampling from a wider image distribution than would be done naturally.

\subsection{Neural network}

In the case of a VAE, the neural network consists of two parts. The \emph{encoder} produces the latent representation from the input, while the \emph{decoder} reconstructs the input from the latent representation. We generalize this terminology and \emph{define} the encoder for an arbitrary RepL method to be that part of the neural network that is used to compute the representation, and the decoder to be the rest of the neural network. Under this definition, the downstream tasks (which could include imitation, classification, reinforcement learning, etc.) only require the encoder, not the decoder. Note that the ``decoder'' may not convert the learned representation into some human-interpretable format; it is simply those parts of the neural network that are required by the RepL method but that do not serve to compute the representation.

\subsubsection{Encoder}
The encoder is the core component of a representation learner: it is responsible for mapping input targets $x$ into $z$ vectors that are used as the learnt representation in downstream tasks. 

\prg{Recurrent encoders.} In some cases, a ``context'' could be a sequence of frames instead of a single frame, and the encoder could compress that into a single representation of the past. This paper doesn't address recurrent encoders, opting instead to make all encoders operate on single framestacks. 

\prg{Momentum encoders.} In contrastive tasks, learning a high-quality representation often requires large batch sizes, since the difficulty of the contrastive task scales with the number of negatives. However, batches of the appropriate difficulty can be so large that encoding the negative targets becomes prohibitively compute- and memory-intensive. \citet{he2020momentum} propose reusing negative targets from previous batches to alleviate this cost. One challenge with reusing targets is that the encoder can change too quickly during training, in which case negative targets from previous batches become ``stale''. 
Thus \citet{he2020momentum} use a separate \textit{target encoder} which is updated slowly enough that targets do not become stale too quickly. Specifically, the target encoder's weights $\theta_t$ are updated to track the main context encoder weights $\theta_c$ using the update rule $\theta_t \leftarrow \alpha \theta_t + (1 - \alpha) \theta_c$. $\alpha$ is referred to as a \textit{momentum} parameter, and is typically set to some value close to 1 (e.g. $\alpha = 0.999$).

\subsubsection{Decoder} 
Decoders are optional neural network layers applied before a loss is calculated, but which are \emph{not} included in the learnt encoder used at transfer time. They take in the $z$ output by the encoder (and optional additional information), and produce an input to the loss function. 

\prg{Image reconstruction.} 
The most common historical form of decoder in a RepL algorithm is the image reconstruction decoder, which has historically been used by VAEs and similar model designs to ``decode'' a predicted image from a representation bottleneck. This predicted image is used in calculating a MLE loss against the true image, but is discarded before downstream transfer tasks. 

\prg{Projection heads.} Projection heads are multi-layer perceptrons that take in the output of the encoder and project it into a new space over which the loss can then be calculated. Recent work has shown these to be useful for contrastive learning~\cite{chen2020simple}.

\prg{Action conditioning.} Temporal tasks can be made easier by conditioning on the action $a_t$. However, for an encoder to be used for reinforcement learning or imitation, the representation must not depend on the current action $a_t$. Thus, the encoder is only responsible for learning a $z$ representation of the observation $o_t$, and is combined with a representation of the action within the decoder step. 

\subsection{Pretraining vs joint training}
Another question is how to integrate representation learning with an RL algorithm. In image recognition, representation learning is done as a pretraining step. We experiment with this approach in this work, as well as the strategy of "joint training", where we add the representation learning loss as an \emph{auxiliary loss} while performing reinforcement learning.



\section{Experiments}
\label{sec:exp}

Given our framework, it is straightforward to construct RepL algorithms that differ along any of the axes described in \cref{sec:repl-analysis}.
In this section, we create a representative set of such algorithms and evaluate various ways of combining them with imitation learning.
Although some RepL methods appear to be effective on some tasks, we find that the difference between using and not using RepL is often much less than the difference between using and not using augmentations for the imitation policy.
In \cref{sec:discussion}, we discuss possible reasons why RepL does not have a greater effect, and suggest alternative ways that RepL could be used more fruitfully.

\subsection{Experiment setup}

\prg{Environments and training data.} We evaluate on ten tasks taken from three benchmark domains: \dmc{}~\cite{tassa2018deepmind}, Procgen~\cite{cobbe2020leveraging}, and MAGICAL~\cite{toyer2020magical}.
Here we briefly explain our choice of tasks and datasets; for more detailed information (e.g.\ dataset sizes and collection methods), refer to \cref{appendix:hyperparams}.

From \dmc{}, we take image-based versions of the cheetah-run, finger-spin, and reacher-easy tasks.
All three of these are popular benchmark tasks for deep RL and deep IL, and represent a range of difficulties (reacher-easy being the easiest, and cheetah-run being the hardest).
However, they provide limited evaluation of generalisation.
We use a common demonstration set for RepL and IL.

\begingroup
\setlength{\tabcolsep}{6pt}
\begin{table*}[t!]
\caption{Design decisions for representation learning algorithms used in our experiments.}
\label{table:algorithms-main-expt}
\begin{center}
\begin{tabular}{@{}lccccc@{}}
\toprule
{\bf Algorithm}  & {\bf Task}          & {\bf Context}  & {\bf Target} & {\bf Act} & {\bf Aug} \\ 
\midrule
Temporal CPC     & Contrastive         & $o_t$          & $o_{t+1}$    & \n & \y \\ 
SimCLR           & Contrastive         & $o_t$          & $o_t$        & \n & \y \\ 
VAE              & Reconstructive      & $o_t$          & $o_t$        & \n & \n \\ 
Dynamics         & Reconstructive      & $o_t, a_t$     & $o_{t+1}$    & \y & \n \\ 
Inverse Dynamics & Reconstructive      & $o_t, o_{t+1}$ & $a_t$        & \n & \n \\ 
\bottomrule
\vspace{-10mm}
\end{tabular}
\end{center}
\end{table*}

From Procgen, we choose the ``easy'' variants of the CoinRun, Fruitbot, Jumper and Miner tasks.
In Procgen, different random initialisations for a given task can have wildly different appearance and structure, but still admit a common optimal policy.
This makes it a much more challenging evaluation of generalization than \dmc{}.
As with \dmc{}, we use the same demonstration set for RepL and IL.

From MAGICAL we choose the MoveToRegion, MoveToCorner, and MatchRegions tasks, which represent a range of difficulty levels (MoveToRegion being the easiest, and MatchRegions being the hardest).
For each task, MAGICAL defines a ``demo variant'' for training and a set of ``test variants'' for evaluating robustness to changes in dynamics, appearance, etc.
Unlike \dmc{} and Procgen, our MAGICAL experiments augment the demonstration set with additional demo variant random rollouts for RepL training.
This models the setting in which it is cheap to collect additional data for self-supervised learning, but expensive to collect demonstrations.
We include more detailed environment setups in Appendix \ref{sec:appendix-experiment}.

\prg{Imitation baselines.} 
Most of our experiments use behavioral cloning (BC)~\cite{pomerleau1991efficient} as the base imitation learning algorithm.
Given a dataset $\mathcal D = \{(x_0, a_0), (x_1, a_1), \ldots\}$ of observation--action tuples drawn from a demonstrator, BC optimises the policy $\pi_\theta(a \mid x)$ to maximise the expected log likelihood,
\[
  \mathcal L(\theta) = \expect{(x,a)\sim\mathcal D}{\log \pi_\theta(a \mid x)}\,.
\]
We combine BC with representation learning in two ways.
First, we use RepL to pretrain all but the final layer of the policy, then fine-tune the policy end-to-end with BC.
This appears to be the most popular approach in the vision literature.
Second, we use RepL as an auxiliary objective during BC training, so that both imitation and representation learning are performed simultaneously.
Importantly, we also do control runs both with and without image augmentations.
The deep RL community has repeatedly found that image augmentations can yield a greater improvement than some sophisticated representation learning  methods~\cite{laskin2020curl,kostrikov2020image}, and so it is important to distinguish between performance gains due to the choice of RepL objective and performance gains due to the use of augmentations.

In addition to BC, we present results with Generative Adversarial Imitation Learning (GAIL)~\cite{ho2016generative} and RepL pretraining.
GAIL treats IL as a game between an imitation policy $\pi_\theta(a \mid x)$ and a discriminator $D_\psi(x,a)$ that must distinguish $\pi_\theta$'s behaviour from that of the demonstrator.
Using alternating gradient descent, GAIL attempts to find a $\theta$ and $\psi$ that attain the saddle point of
\begin{align*}
    \max_\theta \min_\psi \left\{
      - \expect{(x,a)\sim\pi_\theta}{\log D_\psi(x,a)}
      - \expect{(x,a)\sim\mathcal D}{\log(1 - D_\psi(x,a))}
      + w_H H(\pi_\theta)
    \right\}\,.
\end{align*}
Here $H$ is an entropy penalty weighted by regularisation parameter $w_H \geq 0$.
We use augmentations only for the GAIL discriminator, and not the policy (we could not get GAIL to train reliably with policy augmentations).
Discriminator regularisation is of particular interest because past work has shown that discriminator augmentations are essential to obtaining reasonable imitation performance when applying GAIL to image-based environments~\cite{zolna2019task}.
For our experiments combining GAIL with RepL, we use the learned representation to initialize both the GAIL discriminator and the GAIL policy.

\prg{RepL algorithms.} 
Using our modular representation learning framework, we construct five representation learning algorithms described in \cref{table:algorithms-main-expt}. More detailed descriptions are in \cref{sec:appendix-experiment}.

\subsection{Results}

Results are shown in \cref{table:pretrain-bc} for BC + RepL pretraining, and \cref{table:joint-bc} for BC + RepL joint training, and \cref{table:pretrain-gail} for GAIL + RepL pretraining.
Each cell shows mean $\mpm$ standard deviation over at least five random seeds.
We treat IL with augmentations (but no RepL) as our baseline.
We color cells that have a higher mean return than the baseline, and mark them with an asterisk (*) when the difference is significant at $p<0.05$, as measured by a one-sided Welch's t-test without adjustment for multiple comparisons. We include the loss curves for our BC experiments in \cref{app:loss-curves}.
 

\begingroup
\setlength{\tabcolsep}{1pt}
\begin{table*}[t!]
\caption{Pretraining results for BC. We color cells that have a higher mean return than BC with augmentations, and mark them with an asterisk (*) when the difference is significant at $p<0.05$, as measured by a one-sided Welch's t-test without adjustment for multiple comparisons.}
\label{table:pretrain-bc}
\begin{center}
\begin{small}
\begin{tabular}{@{}cccccccccc@{}}
\toprule
\textbf{Env} & \textbf{Task} & \textbf{Dynamics} & \textbf{InvDyn} & \textbf{SimCLR} & \textbf{TemporalCPC} & \textbf{VAE} & \textbf{BC aug} & \textbf{BC no aug} \\
\midrule
\dmc{} & cheetah-run & 482$\mpm$36 & 669$\mpm$18 & 687$\mpm$17 & 661$\mpm$13 & 458$\mpm$39 & 690$\mpm$17 & 617$\mpm$34 & \\
& finger-spin & 718$\mpm$17 & \cellcolor[HTML]{fff7df} 748$\mpm$17* & 726$\mpm$1 & 723$\mpm$4 & \cellcolor[HTML]{fff7df} 751$\mpm$6* & 730$\mpm$9 & \cellcolor[HTML]{fff7df} 940$\mpm$4* & \\
& reacher-easy & 774$\mpm$24 & \cellcolor[HTML]{fff7df} 890$\mpm$14 & \cellcolor[HTML]{fff7df} 907$\mpm$9 & \cellcolor[HTML]{fff7df} 893$\mpm$13 & \cellcolor[HTML]{fff7df} 880$\mpm$20 & 874$\mpm$21 & 452$\mpm$34 & \\ \midrule
Procgen & coinrun-train & 8.1$\mpm$0.4 & 8.0$\mpm$0.2 & 8.0$\mpm$0.5 & \cellcolor[HTML]{fff7df} 8.1$\mpm$0.3 & \cellcolor[HTML]{fff7df} 8.4$\mpm$0.4 & 8.1$\mpm$0.3 & \cellcolor[HTML]{fff7df} 8.7$\mpm$0.6* & \\
& fruitbot-train & 3.2$\mpm$1 & 16.2$\mpm$1.2 & 17.5$\mpm$1.9 & 15.4$\mpm$1.5 & 17.5$\mpm$1.5 & 18.3$\mpm$1.9 & 11.4$\mpm$0.6 & \\
& jumper-train & 8.1$\mpm$0.2 & 8.0$\mpm$0.4 & 7.9$\mpm$0.6 & 7.5$\mpm$0.6 & 7.9$\mpm$0.6 & 8.1$\mpm$1.2 & 7.1$\mpm$1.2 & \\
& miner-train & 4.5$\mpm$1.2 & 5.9$\mpm$0.2 & \cellcolor[HTML]{fff7df} 9.9$\mpm$0.4 & 9.5$\mpm$2.3 & \cellcolor[HTML]{fff7df} 10.4$\mpm$0.3* & 9.8$\mpm$0.3 & 8.1$\mpm$0.3 & \\ \hline
& coinrun-test & 6.3$\mpm$0.8 & \cellcolor[HTML]{fff7df} 6.9$\mpm$0.5 & \cellcolor[HTML]{fff7df} 6.8$\mpm$0.5 & \cellcolor[HTML]{fff7df} 6.8$\mpm$0.4 & \cellcolor[HTML]{fff7df} 7.0$\mpm$0.5 & 6.7$\mpm$0.4 & 6.5$\mpm$0.7 & \\
& fruitbot-test & -3$\mpm$0.9 & \cellcolor[HTML]{fff7df} 15.6$\mpm$1.1 & 13.4$\mpm$1.0 & \cellcolor[HTML]{fff7df} 14.7$\mpm$1.0 & 13.2$\mpm$1.0 & 13.7$\mpm$1.1 & 2.2$\mpm$0.6 & \\
& jumper-test & 3.2$\mpm$0.4 & 3.9$\mpm$0.3 & 3.6$\mpm$0.4 & 3.7$\mpm$0.5 & 3.4$\mpm$0.5 & 3.9$\mpm$0.5 & \cellcolor[HTML]{fff7df} 4.6$\mpm$0.4 & \\
& miner-test & 0.6$\mpm$0.1 & 2.6$\mpm$0.1 & 2.6$\mpm$0.4 & \cellcolor[HTML]{fff7df} 3.1$\mpm$0.4 & \cellcolor[HTML]{fff7df} 2.7$\mpm$0.3 & 2.7$\mpm$0.4 & 0.8$\mpm$0.1 & \\
\midrule
MAGI- & MatchRegions & 0.42$\mpm$0.04 & 0.42$\mpm$0.04 & 0.42$\mpm$0.03 & 0.41$\mpm$0.01 & 0.42$\mpm$0.03 & 0.43$\mpm$0.02 & 0.28$\mpm$0.08 \\
CAL & MoveToCorner & \cellcolor[HTML]{fff7df}0.84$\mpm$0.07 & \cellcolor[HTML]{fff7df}0.83$\mpm$0.04 & \cellcolor[HTML]{fff7df}0.83$\mpm$0.04* & \cellcolor[HTML]{fff7df}0.80$\mpm$0.02 & \cellcolor[HTML]{fff7df}0.78$\mpm$0.06 & 0.78$\mpm$0.05 & 0.72$\mpm$0.04 \\
& MoveToRegion & \cellcolor[HTML]{fff7df}0.82$\mpm$0.02* & \cellcolor[HTML]{fff7df}0.83$\mpm$0.02* & \cellcolor[HTML]{fff7df}0.82$\mpm$0.01* & \cellcolor[HTML]{fff7df}0.81$\mpm$0.01* & \cellcolor[HTML]{fff7df}0.81$\mpm$0.05* & 0.74$\mpm$0.02 & \cellcolor[HTML]{fff7df}0.81$\mpm$0.04* \\
\bottomrule
\end{tabular}
\vspace{-5mm}
\end{small}
\end{center}
\end{table*}
\endgroup

\prg{BC pretraining results.}
In the pretraining setting, we see that none of our RepL algorithms consistently yield large improvements across all (or even most) tasks.
Indeed, the relative impact of adding representation learning tends to be lower than the impact of adding or removing augmentations.
Although adding augmentations to BC usually yields a large improvement, there are a handful of tasks where adding augmentations substantially decreases performance; we remark further on this below.
Note that most of our RepL algorithms do seem to yield an improvement in MoveToRegion, suggesting that there may still be value to RepL for a narrower set of tasks and datasets.

\prg{BC joint training results.}
When using joint training as an auxiliary loss, we similarly see that no one RepL method consistently improves performance across all benchmark tasks.
However, in the \dmc{} tasks, we do see consistent improvement over the baseline for all RepL methods.
This suggests that our RepL methods provide benefit in some environments, but are sensitive to the choice of task.

\begingroup
\setlength{\tabcolsep}{1pt}
\begin{table*}[t!]
\caption{Joint training results for BC. We color cells that have a higher mean return than BC with augmentations, and mark them with an asterisk (*) when the difference is significant at $p<0.05$, as measured by a one-sided Welch's t-test without adjustment for multiple comparisons.}
\label{table:joint-bc}
\begin{center}
\begin{small}
\begin{tabular}{@{}cccccccccc@{}}
\toprule
\textbf{Env} & \textbf{Task} & \textbf{Dynamics} & \textbf{InvDyn} & \textbf{SimCLR} & \textbf{TemporalCPC} & \textbf{VAE} & \textbf{BC aug} & \textbf{BC no aug} \\
\midrule
\dmc{} & cheetah-run & \cellcolor[HTML]{fff7df} 723$\mpm$14* & \cellcolor[HTML]{fff7df} 716$\mpm$23* & \cellcolor[HTML]{fff7df} 717$\mpm$11* & \cellcolor[HTML]{fff7df} 716$\mpm$16* & \cellcolor[HTML]{fff7df} 724$\mpm$12* & 690$\mpm$17 & 617$\mpm$34 & \\
& finger-spin & \cellcolor[HTML]{fff7df} 755$\mpm$6* & \cellcolor[HTML]{fff7df} 755$\mpm$12* & \cellcolor[HTML]{fff7df} 732$\mpm$15 & 725$\mpm$12 & \cellcolor[HTML]{fff7df} 755$\mpm$3* & 730$\mpm$9 & \cellcolor[HTML]{fff7df} 940$\mpm$4* & \\
& reacher-easy & \cellcolor[HTML]{fff7df} 898$\mpm$19 & \cellcolor[HTML]{fff7df} 903$\mpm$10* & \cellcolor[HTML]{fff7df} 889$\mpm$14 & \cellcolor[HTML]{fff7df} 912$\mpm$18* & \cellcolor[HTML]{fff7df} 903$\mpm$8* & 874$\mpm$21 & 452$\mpm$34 & \\ \midrule
Proc- & coinrun-train & 8.0$\mpm$0.4 & 7.1$\mpm$0.3 & 8.0$\mpm$0.5 & \cellcolor[HTML]{fff7df} 8.6$\mpm$0.5* & 7.9$\mpm$0.2 & 8.1$\mpm$0.3 & \cellcolor[HTML]{fff7df} 8.7$\mpm$0.6* & \\
gen & fruitbot-train & 17.0$\mpm$0.7 & 6.6$\mpm$1.4 & 13.4$\mpm$1.9 & 11.4$\mpm$0.7 & 15.4$\mpm$1.0 & 18.3$\mpm$1.9 & 11.4$\mpm$0.6 & \\
& jumper-train & 7.9$\mpm$0.5 & 8.1$\mpm$0.4 & 8.0$\mpm$0.4 & 8.0$\mpm$0.3 & \cellcolor[HTML]{fff7df} 8.3$\mpm$0.5 & 8.1$\mpm$1.2 & 7.1$\mpm$1.2 & \\
& miner-train & 8.9$\mpm$0.8 & 8.9$\mpm$0.7 & 8.7$\mpm$0.3 & 7.1$\mpm$0.8 & 8.6$\mpm$0.7 & 9.8$\mpm$0.3 & 8.1$\mpm$0.3 & \\ \hline
& coinrun-test & 6.4$\mpm$0.4 & 6.0$\mpm$0.5 & 6.6$\mpm$0.3 & 6.2$\mpm$0.5 & \cellcolor[HTML]{fff7df} 6.9$\mpm$0.4 & 6.7$\mpm$0.4 & 6.5$\mpm$0.7 & \\
& fruitbot-test & 10.9$\mpm$0.7 & 3.3$\mpm$1.1 & 8.5$\mpm$1.5 & 6.4$\mpm$1.2 & 10.4$\mpm$1.6 & 13.7$\mpm$1.1 & 2.2$\mpm$0.6 & \\
& jumper-test & 3.4$\mpm$0.3 & \cellcolor[HTML]{fff7df} 4.8$\mpm$0.2* & 3.8$\mpm$0.3 & 3.4$\mpm$0.3 & 3.9$\mpm$0.7 & 3.9$\mpm$0.5 & \cellcolor[HTML]{fff7df} 4.6$\mpm$0.4* & \\
& miner-test & 2.0$\mpm$0.2 & 1.9$\mpm$0.3 & 1.8$\mpm$0.3 & 1.0$\mpm$0.2 & 2.0$\mpm$0.3 & 2.7$\mpm$0.4 & 0.8$\mpm$0.1 & \\
\midrule
MAGI- & MatchRegions & \cellcolor[HTML]{fff7df}0.44$\mpm$0.02 & 0.23$\mpm$0.08 & 0.41$\mpm$0.02 & 0.01$\mpm$0.01 & 0.41$\mpm$0.03 & 0.43$\mpm$0.03 & 0.31$\mpm$0.02 \\
CAL & MoveToCorner & 0.78$\mpm$0.07 & 0.30$\mpm$0.22 & 0.76$\mpm$0.05 & 0.02$\mpm$0.02 & \cellcolor[HTML]{fff7df}0.82$\mpm$0.06 & 0.80$\mpm$0.05 & 0.70$\mpm$0.09 \\
& MoveToRegion & \cellcolor[HTML]{fff7df}0.76$\mpm$0.02 & 0.35$\mpm$0.24 & 0.74$\mpm$0.01 & 0.47$\mpm$0.07 & \cellcolor[HTML]{fff7df}0.77$\mpm$0.02 & 0.75$\mpm$0.02 & \cellcolor[HTML]{fff7df}0.78$\mpm$0.04 \\
\bottomrule
\end{tabular}
\vspace{-5mm}
\end{small}
\end{center}
\end{table*}
\endgroup

\prg{Effect of augmentations on BC.}
Incorporating augmentations into BC training tended to yield the largest effect of any technique considered in this work, even without an explicit representation learning loss. 
In roughly half of the environments studied, this had a substantial impact on reward, and reward increased $150\%$ or more in reacher-easy, Fruitbot, and MatchRegions. 
However, environments seem to be bimodal in their response to augmentations: in a handful of environments (finger-spin, coinrun-train, jumper-test, and MoveToRegion), adding augmentations leads to consistently \emph{worse} performance. This effect is particularly dramatic in finger-spin, which we believe is a result of the fact that relevant objects in the environment always stay fixed.
Consequently, translational augmentations don't aid generalization, and rotational augmentations can be confused with true signal (since the angle of the finger determines the ideal action). 
Because augmentation already yields large benefits, many of the representation learning algorithms do not provide much additional gain on top of BC-Augs, even when they perform substantially better than BC-NoAugs. 
This result is consistent with the finding by \citet{laskin2020reinforcement} that simply augmenting input frames in reinforcement learning produced performance on par with sophisticated representation learning methods.

\prg{GAIL pretraining results.} GAIL pretraining results mirror those for BC pretraining, but with even fewer statistically significant deviations from baseline performance.
We see that augmentation can be even more important for GAIL than it is for BC.
For instance, GAIL with discriminator augmentations obtains higher return on finger-spin than BC does, but obtains a return of 0 when discriminator augmentations are removed.
This is consistent with the observation of \citeauthor{zolna2019task} that strict regularisation is essential to make GAIL perform well in image-based domains~\cite{zolna2019task}.

\begingroup
\setlength{\tabcolsep}{1pt}
\begin{table*}[t!]
\caption{Pretraining results for GAIL. We color cells that have a higher mean return than BC with augmentations, and mark them with an asterisk (*) when the difference is significant at $p<0.05$, as measured by a one-sided Welch's t-test without adjustment for multiple comparisons.
For the sake of space, we abbreviate TemporalCPC to $t$CPC.}
\label{table:pretrain-gail}
\begin{center}
\begin{small}
\begin{tabular}{@{}ccccccccccc@{}}
\toprule
\textbf{Env} & \textbf{Task} & \textbf{Dynamics} & \textbf{InvDyn} & \textbf{SimCLR} & \textbf{$t$CPC} & \textbf{VAE} & \textbf{GAIL aug} & \textbf{GAIL no aug} \\
\midrule
\dmc{} & cheetah-run & 380$\mpm$76 & 320$\mpm$61 & 265$\mpm$58 & 360$\mpm$74 & 375$\mpm$33 & 449$\mpm$67 & 75$\mpm$40 \\
& finger-spin& \cellcolor[HTML]{fff7df}868$\mpm$14 & \cellcolor[HTML]{fff7df}886$\mpm$8* & 800$\mpm$23 & 748$\mpm$72 & \cellcolor[HTML]{fff7df}868$\mpm$18 & 868$\mpm$12 & 0$\mpm$0 \\
& reacher-easy & 53$\mpm$24 & 73$\mpm$51 & 21$\mpm$23 & 118$\mpm$88 & 122$\mpm$89 & 221$\mpm$162 & 89$\mpm$88 \\
\midrule
Proc- & coinrun-train & \cellcolor[HTML]{fff7df}5.9$\mpm$0.29* & \cellcolor[HTML]{fff7df}5.85$\mpm$0.51* & 2.15$\mpm$1.53 & 3.28$\mpm$2.62 & \cellcolor[HTML]{fff7df}3.54$\mpm$1.22 & 3.31$\mpm$0.44 & 2.80$\mpm$0.89 \\
gen & fruitbot-train & -2.81$\mpm$0.1 & \cellcolor[HTML]{fff7df}-2.37$\mpm$0.55 & -2.47$\mpm$0.15 & \cellcolor[HTML]{fff7df}-2.38$\mpm$0.31 & -2.49$\mpm$0.22 & -2.42$\mpm$0.42 & -2.63$\mpm$0.30 \\
& jumper-train & 3.31$\mpm$0.31 & 3.17$\mpm$0.40 & 3.36$\mpm$0.53 & 2.69$\mpm$1.31 & \cellcolor[HTML]{fff7df}3.45$\mpm$0.70 & 3.44$\mpm$0.52 & \cellcolor[HTML]{fff7df}3.47$\mpm$0.53 \\
& miner-train & 0.53$\mpm$0.12 & 0.60$\mpm$0.11 & 0.53$\mpm$0.14 & \cellcolor[HTML]{fff7df}0.84$\mpm$0.14* & 0.51$\mpm$0.07 & 0.65$\mpm$0.10 & \cellcolor[HTML]{fff7df}0.77$\mpm$0.18 \\
\hline
& coinrun-test & \cellcolor[HTML]{fff7df}6.1$\mpm$0.9* & \cellcolor[HTML]{fff7df}5.91$\mpm$0.16* & 2.11$\mpm$1.61 & 3.35$\mpm$2.74 & 3.01$\mpm$1.10 & 3.44$\mpm$0.68 & 2.77$\mpm$0.84 \\
& fruitbot-test & -2.44$\mpm$0.49 & -2.65$\mpm$0.24 & -2.55$\mpm$0.30 & -2.65$\mpm$0.14 & -2.85$\mpm$0.33 & -2.44$\mpm$0.50 & -2.51$\mpm$0.44 \\
& jumper-test & 2.56$\mpm$0.52 & 2.53$\mpm$0.64 & 3.15$\mpm$0.45 & 2.35$\mpm$0.81 & 2.75$\mpm$0.59 & 3.25$\mpm$0.42 & 3.15$\mpm$0.20 \\
& miner-test & 0.36$\mpm$0.04 & 0.57$\mpm$0.07 & 0.55$\mpm$0.24 & \cellcolor[HTML]{fff7df}0.87$\mpm$0.15* & 0.50$\mpm$0.17 & 0.65$\mpm$0.17 & \cellcolor[HTML]{fff7df}0.66$\mpm$0.14 \\
\midrule
MAGI- & MatchRegions & 0.42$\mpm$0.10 & 0.34$\mpm$0.12 & \cellcolor[HTML]{fff7df}0.47$\mpm$0.04 & 0.39$\mpm$0.12 & 0.30$\mpm$0.15 & 0.46$\mpm$0.06 & 0.22$\mpm$0.12 \\
CAL & MoveToCorner & 0.48$\mpm$0.09 & 0.45$\mpm$0.10 & \cellcolor[HTML]{fff7df}0.52$\mpm$0.07 & \cellcolor[HTML]{fff7df}0.55$\mpm$0.15 & \cellcolor[HTML]{fff7df}0.62$\mpm$0.11* & 0.49$\mpm$0.08 & \cellcolor[HTML]{fff7df}0.55$\mpm$0.14 \\
& MoveToRegion & 0.72$\mpm$0.07 & 0.74$\mpm$0.04 & 0.74$\mpm$0.06 & \cellcolor[HTML]{fff7df}0.76$\mpm$0.03 & 0.75$\mpm$0.07 & 0.75$\mpm$0.09 & 0.60$\mpm$0.14 \\
\bottomrule
\end{tabular}
\vspace{-7mm}
\end{small}
\end{center}
\end{table*}
\endgroup

\section{Discussion \& future work}\label{sec:discussion}


\prg{Contrasting image classification and imitation learning datasets.}
The use of self-supervised representation learning for pretraining has met with notable success in image classification~\cite{chen2020simple}. By comparison, results from RL literature have been mixed, with some positive results, but also several works~\cite{laskin2020reinforcement,kostrikov2020image} which claim that RepL adds little value relative to image augmentation---a result which we observe in imitation as well. Given this, it's natural to wonder \emph{why} successes from supervised learning have not been reproduced in sequential decision making problems such as RL and imitation.

\begin{wrapfigure}{r}{0.7\textwidth}
\begin{center}
  \vspace{-22pt}
  \includegraphics[width=0.65\textwidth]{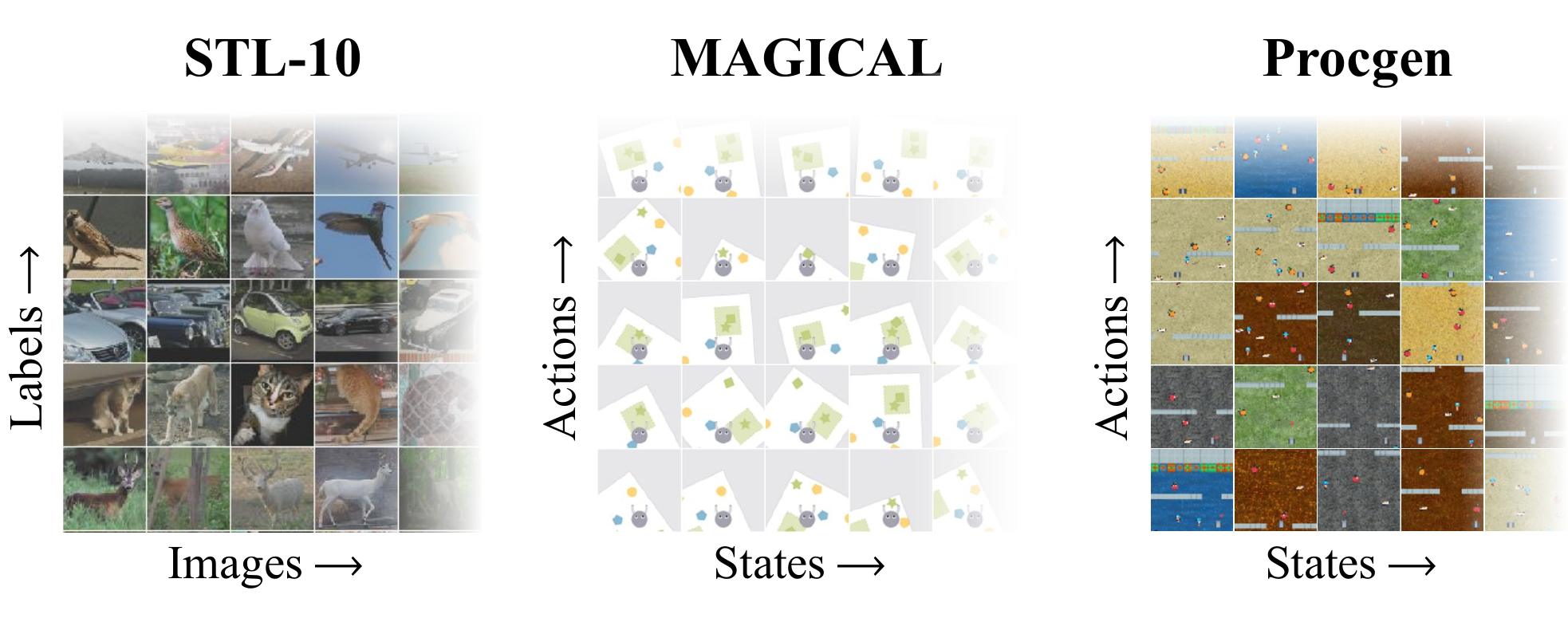}
  \vspace{-2mm}
  \caption{We show a sample of STL-10, MAGICAL, and Procgen images. Images on the same row have the same label (bird, car, etc.) or expert action (up, down, etc.). It can be easier to tell whether two images have the same label in classification than in IL tasks.}
\label{fig:clss-vs-seq}
\vspace{-5mm}
\end{center}
\end{wrapfigure}

The case of Behavioural Cloning (BC) is particularly illustrative.
BC uses the same optimization algorithms, loss types, and network architectures as other forms of image classification, so if RepL is less helpful for BC than for other forms of classification then it must be due to the choice of training and evaluation data.
For the sake of illustrating differences in data distributions, \cref{fig:clss-vs-seq} compares the STL-10 dataset---a typical image classification task---with datasets for MAGICAL and Procgen. dm\_control is not pictured because it has a continuous action space, so there was not a natural separation of images by action along the $y$ axis.


One notable difference in \cref{fig:clss-vs-seq} is that there is less between-class variation in MAGICAL and Procgen than in STL-10: the choice of action is often influenced by fine-grained, local cues in the environment, rather than the most visually salient axes of variation (background, mean color, etc.).
For example, in MAGICAL the sets of states that correspond to the ``forward'' and ``left'' demonstrator actions cover a similar visual range.
Indeed, the agent's choice between ``left'' and ``right'' could change if its heading shifted by just a few degrees, even though this visual change would not be obvious to a human.
In contrast, STL-10 exhibits substantial between-class variation: it's hard to confuse a the sky-blue background and metal texture of a plane for the natural setting and fur of a deer.
Thus, a RepL method that simply captures the most visually salient differences between classes may be much more useful for classification on STL-10 than for control on MAGICAL or Procgen.

\begin{figure}[]
    \begin{subfigure}{0.3\textwidth}
        \centering 
        \includegraphics[width=\textwidth]{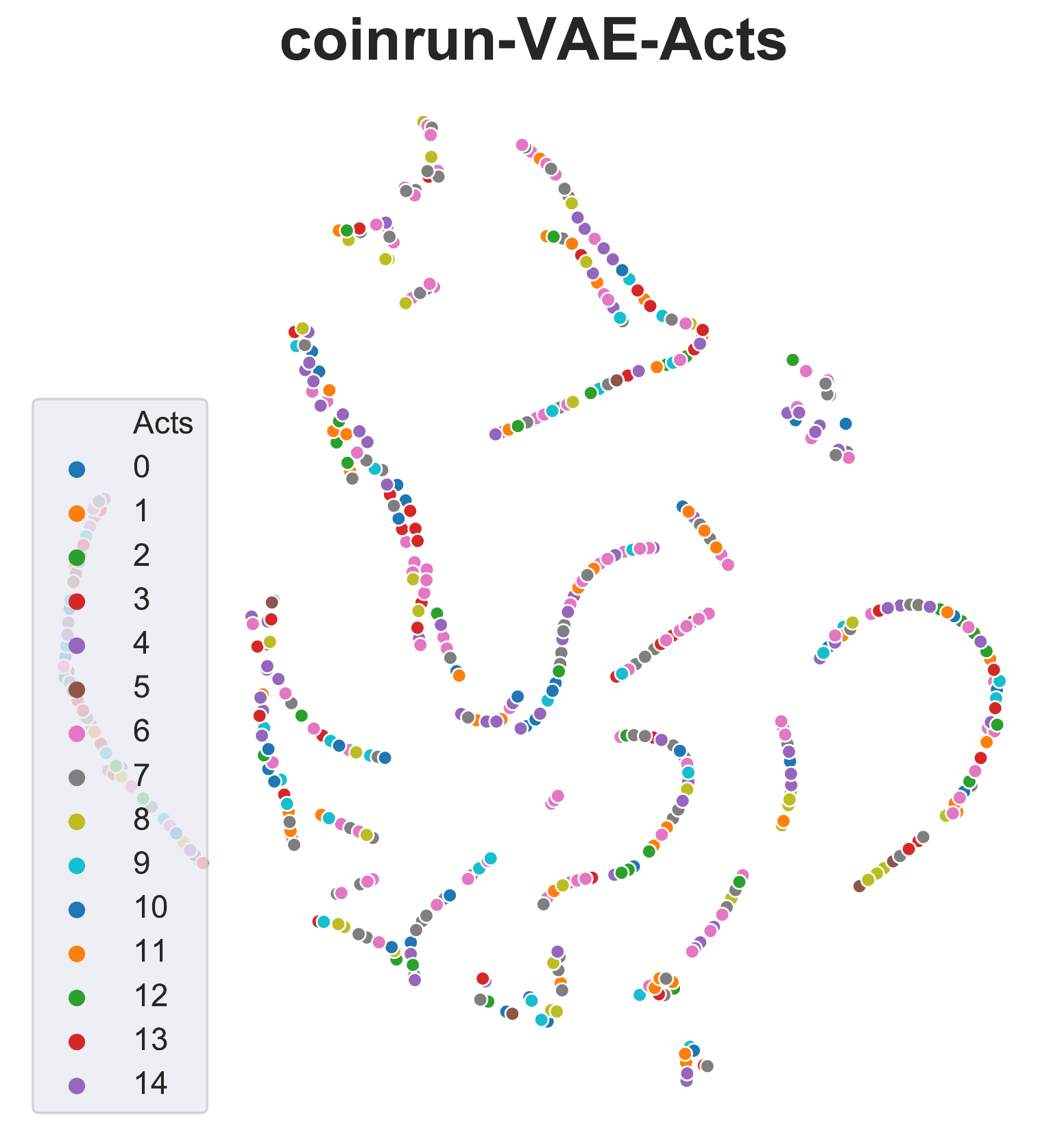}
    \end{subfigure}
    \begin{subfigure}{0.3\textwidth}
        \centering 
        \includegraphics[width=\textwidth]{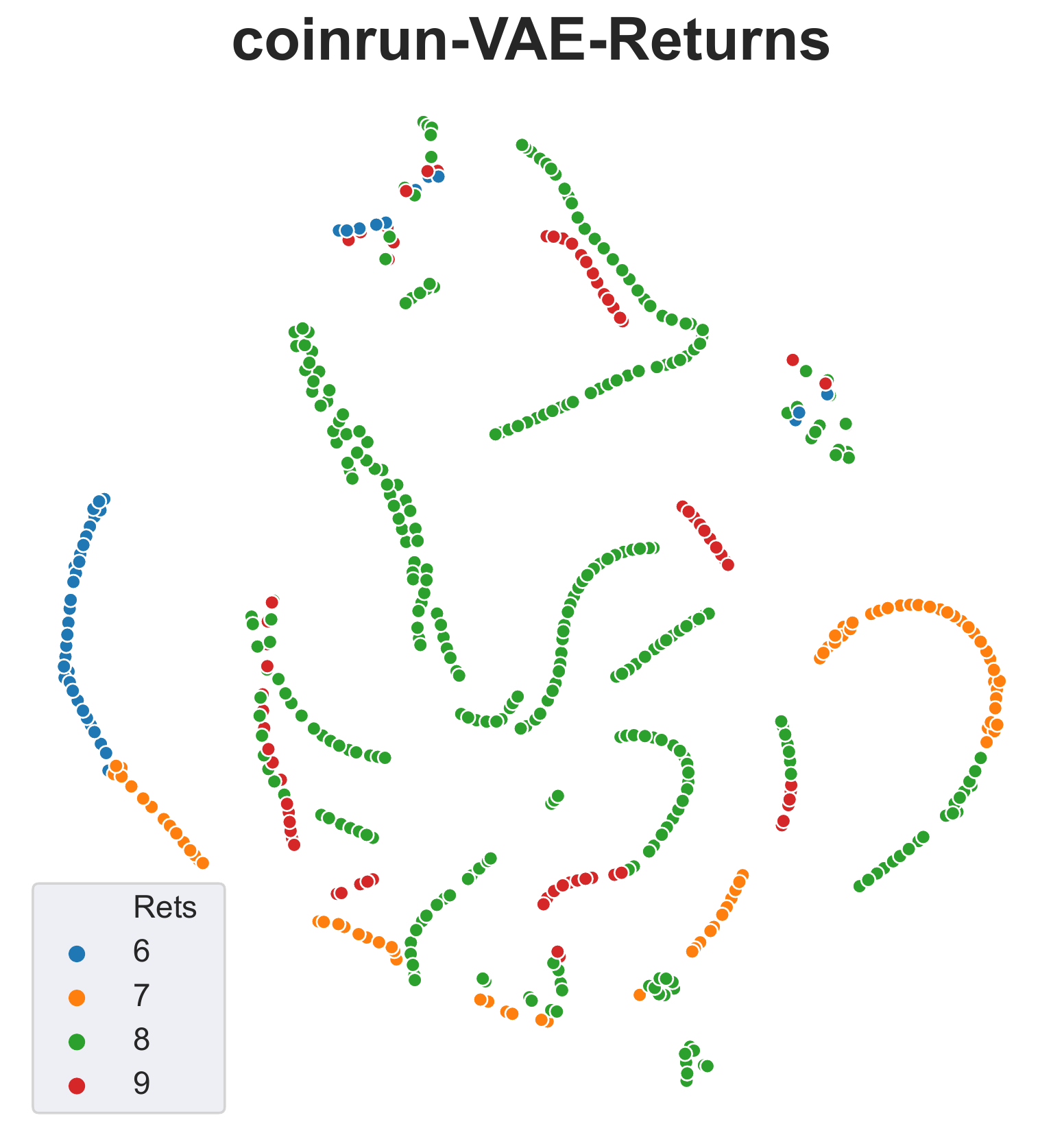}
    \end{subfigure}
    \begin{subfigure}{0.3\textwidth}
        \centering
        \includegraphics[width=\textwidth]{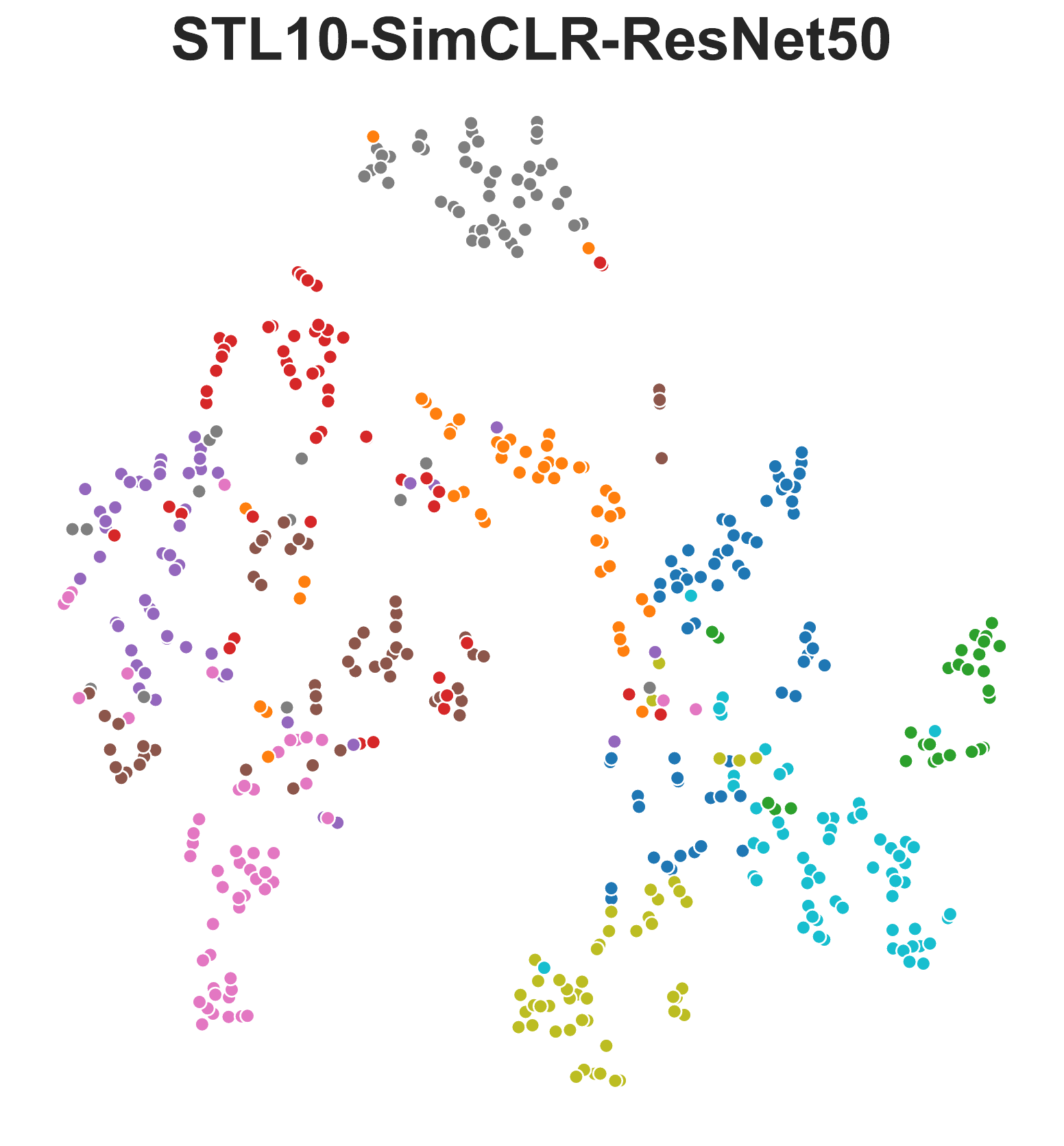}
    \end{subfigure}
    \caption{t-SNE embedding of representations from a VAE encoder on CoinRun, labeled with the corresponding actions (left) and discretized returns (middle).
    Returns are estimated by applying GAE to an expert PPO demonstrator, then discretized by rounding to the nearest whole number to produce a ``label''.
    We compare these with STL-10 image representations generated by a ResNet50 pretrained with SimCLR, colored by class (right).}
\label{fig:cluster-analysis}
\vspace{-5mm}
\end{figure}

\prg{What is the right downstream prediction target?}
Both GAIL and BC attempt to learn a policy that predicts expert actions from observations.
We've argued that RepL algorithms may be focusing primarily on the most visually salient differences between states, at the expense of the fine-grained features necessary for action prediction.
However, it could be that reward- and value-prediction benefit more from a representation that captures mostly coarse-grained visual differences.
Moreover, \citeauthor{yang2021representation}~\cite{yang2021representation} have observed that state-based (as opposed to image-based) offline Q-learning \textit{does} benefit from existing RepL techniques, even though state-based BC does not.
Together, these facts suggest existing RepL methods might be more helpful when the downstream prediction target is value or reward rather than action.

To explore this hypothesis, we visualize how well RepL-learned representations align with action labels, estimated expert returns, and trajectory IDs.
In Figure~\ref{fig:cluster-analysis} we show t-SNE projections of observation embeddings taken from seven expert CoinRun demonstrations.
The embeddings were generated by a VAE-pretrained encoder.
We compare these with t-SNE clusters generated from a ResNet50 with SimCLR on ImageNet, then evaluated on STL-10 (a resized ImageNet subset).

Representations from a well-trained encoder should cluster nicely according to the label (e.g. classes, actions) used for the downstream task.
We see this with the STL-10 embeddings, which cluster nicely by class.
In contrast, we see that our encoders for CoinRun do not produce embeddings that cluster nicely by action.
However, they do seem to cluster readily by estimated expert returns. 
This is likely a consequence of the events that cause states to have high value---such as being close to the far wall with the coin---depend primarily on coarse-grained features of the state.
We speculate that this is likely true in MAGICAL, too, where the reward function tends to depend only on salient features like whether the agent is overlapping with any of the colored goal regions.

Our negative results for GAIL and RepL provide reason to be cautious about our conjecture that reward functions (and value functions) are more amenable to RepL.
A GAIL discriminator is similar to a reward function, but the overall performance of GAIL does not change much when pretraining the discriminator with RepL.
On the other hand, it is worth noting that the GAIL discriminator does not in general converge to a valid reward function for the task, so this is not a direct test of the hypothesis that reward learning is more amenable to RepL pretraining than policy learning.
We therefore believe it is still worth investigating whether imitation learning algorithms that directly learn reward functions~\cite{fu2017learning} or value functions~\cite{reddy2019sqil} benefit more from RepL than algorithms that learn policies.

\prg{The importance of using diverse benchmark tasks.}
Our experiment results in \cref{table:joint-bc} showed much greater benefit for RepL on \dmc{} than on Procgen and MAGICAL. This underscores the importance of evaluating across multiple benchmarks: had we only used \dmc, we might have erroneously concluded that RepL is typically helpful for BC.

The finger-spin (DMC) and CoinRun (Procgen) tasks provide a useful illustration of how differences in performance across tasks can arise.
\cref{fig:dmc-vs-procgen} shows example saliency maps~\cite{simonyan2013deep} generated by SimCLR-pretrained encoders in these two tasks.
In finger-spin, the SimCLR encoder mostly attends to foreground objects, while in CoinRun it attends to the background.
This makes sense: the boundary between the background and terrain is easy to detect and shifts rapidly as the agent moves, so paying attention to the shape of background is quite helpful for distinguishing between frames.
Unfortunately, semantically important foreground features in CoinRun, such as obstacles and gold, are less discriminative, which is why we believe SimCLR is not dedicating as much model capacity to them.
In contrast, the background in finger-spin changes very little, so SimCLR is forced to attend to foreground objects that change position between frames. 

More generally, we believe that differences between RepL performance across tasks are due to implicit assumptions that our (unsupervised) RepL algorithms make about what kinds of features are important.
For tasks that do not match these assumptions, the representation learning algorithms will do poorly, regardless of how much data is available.
In our SimCLR example, information about background shapes crowds out task-relevant cues like the distance between the agent and an obstacle.
It is therefore important for future research to (1) consider whether the implicit assumptions underlying a given RepL algorithm are likely to help models acquire useful invariances for the desired tasks; and (2) test on multiple domains to ensure that the claimed improvements are robust across environments.

\begin{figure}[]
    \centering 
    \includegraphics[width=0.9\textwidth]{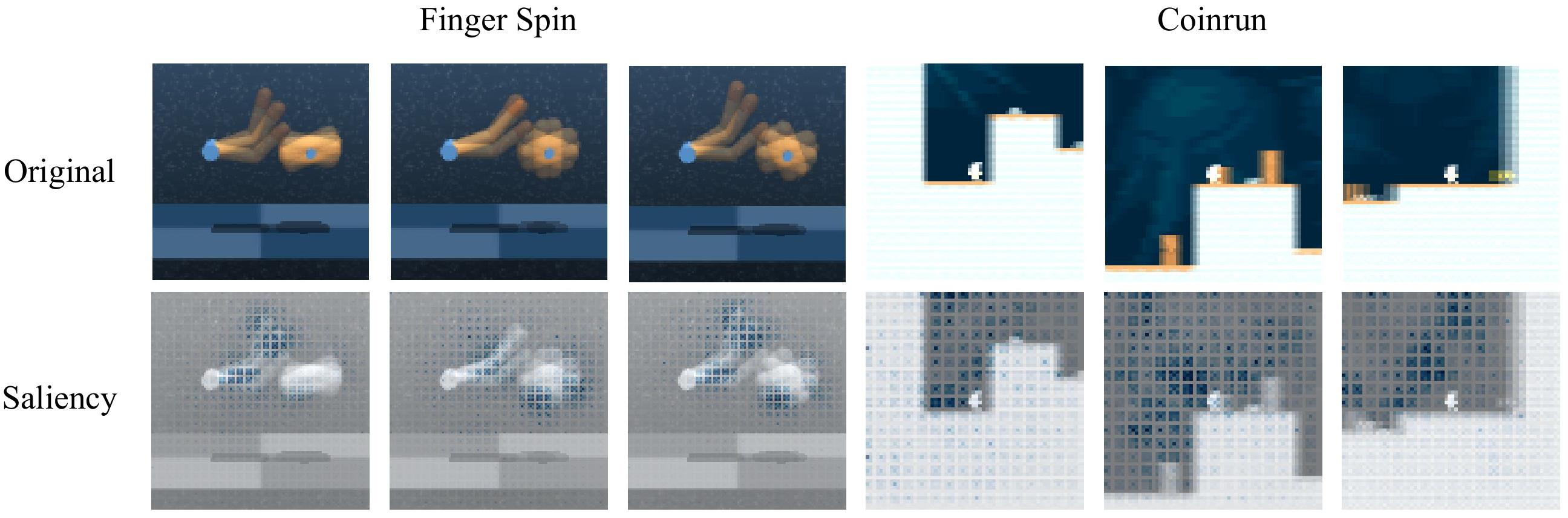}
     \caption{Saliency map generated by an encoder trained using SimCLR.
     Top row shows input frames, averaged across a three-frame stack of inputs.
     Bottom row shows saliency map overlayed on top of grayscale images, with darker blue shading indicating greater influence over the network's output.
     Notice that SimCLR attends mainly to the foreground in DMC, and mainly to the background in CoinRun.}
\label{fig:dmc-vs-procgen}
\vspace{-5mm}
\end{figure}

\section{Conclusion}

We have seen that, when compared against a well-tuned IL baseline using image augmentations, the impacts of representation learning for imitation are limited. On some benchmark suites it appears that it helps, while on others there is not much impact, suggesting that the effect of RepL is quite benchmark-specific. Our analysis has identified several hypotheses that could help understand \emph{when} and \emph{where} representation learning can be useful. We are excited to see future work investigate these hypotheses, and hope the EIRLI framework can serve as a useful starting point for any such investigations.



\begin{ack}
The authors would like to thank Michael Chang, Ben Eysenbach, Aravind Srinivas, and Olivia Watkins for feedback on earlier versions of this work.
This work was supported in part by the DOE CSGF under grant number DE-SC0020347, along with a grant from the Open Philanthropy Project and computational support from Google.
\end{ack}

\bibliography{references}
\bibliographystyle{unsrtnat}

\newpage
\section*{Checklist}


\begin{enumerate}

\item For all authors...
\begin{enumerate}
  \item Do the main claims made in the abstract and introduction accurately reflect the paper's contributions and scope?
    \answerYes{}
  \item Did you describe the limitations of your work?
    \answerYes{}
  \item Did you discuss any potential negative societal impacts of your work?
    \answerYes{} Both the limitations and potential societal impacts are discussed in the appendix.
  \item Have you read the ethics review guidelines and ensured that your paper conforms to them?
    \answerYes{}
\end{enumerate}

\item If you are including theoretical results...
\begin{enumerate}
  \item Did you state the full set of assumptions of all theoretical results?
    \answerNA{}
	\item Did you include complete proofs of all theoretical results?
    \answerNA{}
\end{enumerate}

\item If you ran experiments (e.g. for benchmarks)...
\begin{enumerate}
  \item Did you include the code, data, and instructions needed to reproduce the main experimental results (either in the supplemental material or as a URL)?
    \answerYes{}
  \item Did you specify all the training details (e.g., data splits, hyperparameters, how they were chosen)?
    \answerYes{}
	\item Did you report error bars (e.g., with respect to the random seed after running experiments multiple times)?
    \answerYes{}
	\item Did you include the total amount of compute and the type of resources used (e.g., type of GPUs, internal cluster, or cloud provider)?
    \answerYes{} The information is provided in the Appendix.
\end{enumerate}

\item If you are using existing assets (e.g., code, data, models) or curating/releasing new assets...
\begin{enumerate}
  \item If your work uses existing assets, did you cite the creators?
    \answerYes{}
  \item Did you mention the license of the assets?
    \answerYes{} We use the MIT licence for our codebase, discussed in the Appendix. 
  \item Did you include any new assets either in the supplemental material or as a URL?
    \answerYes{}
  \item Did you discuss whether and how consent was obtained from people whose data you're using/curating?
    \answerNA{}
  \item Did you discuss whether the data you are using/curating contains personally identifiable information or offensive content?
    \answerNA{} 
\end{enumerate}

\item If you used crowdsourcing or conducted research with human subjects...
\begin{enumerate}
  \item Did you include the full text of instructions given to participants and screenshots, if applicable?
    \answerNA{}
  \item Did you describe any potential participant risks, with links to Institutional Review Board (IRB) approvals, if applicable?
    \answerNA{}
  \item Did you include the estimated hourly wage paid to participants and the total amount spent on participant compensation?
    \answerNA{}
\end{enumerate}

\end{enumerate}

\newpage
\appendix
\setcounter{table}{0}
\renewcommand{\thetable}{A\arabic{table}}

\section{Design choices for reinforcement learning}
\begingroup
\setlength{\tabcolsep}{2pt} 
\begin{table}[H]
\caption{Design choices made in representation learning for reinforcement learning. Act, Aug, Mom, Proj and Comp respectively show whether action conditioning, augmentation, momentum, projection heads, and compression were used. P/J determines whether the representation learning is an initial (P)retraining step, or is (J)ointly learned alongside reinforcement learning. R/C/B/N in the Task column refer to Reconstruction, Contrastive, Bootstrap, or None. Note that different papers may use different sets of augmentations.
}
\label{table:existing-algorithms-rl}
\begin{center}
\begin{tabular}{@{}lcccccccccc@{}}
\toprule
{\bf Algorithm} & {\bf Task}    & {\bf RL alg.}  & {\bf Context} & {\bf Target}          & {\bf Act} & {\bf Aug} & {\bf Mom} & {\bf Proj} & {\bf Comp} & {\bf P/J} \\
\midrule
World models~\cite{ha2018world}    & R   & CMA-ES    & $o_t$         & $o_t, o_{t+1}$        & \y  & \n  & \n  & \n  & \n  & P \\
DVRL~\cite{igl2018deep}     & R   & A2C         & $o_t$         & $o_{t+k}$             & \y  & \n   & \n  & \n  & \n  & J \\
PlaNet~\cite{hafner2019learning}          & R   & MPC + CEM & $o_{1:t}$     & $o_{t+1:T}, r_{t+1:T}$& \y  & \n  & \n  & \n  & \n  & J \\
SLAC~\cite{lee2019stochastic}     & R   & SAC         & $o_t$         & $o_{t+1}$             & \y  & \n   & \n  & \n  & \n  & J \\
Poke~\cite{agrawal2016learning}            & R         & -       & $o_t, o_{t+1}$ & $a_t$        & - & \n & \n & \n & \n & - \\
RAD~\cite{laskin2020reinforcement}             & N          & PPO, SAC  & $o_t$         & -                   & - & \y  & - & - & \n  & J \\
DrQ~\cite{kostrikov2020image}             & N          & DQN, SAC  & $o_t$         & -                   & - & \y  & - & - & \n  & J \\
CURL~\cite{laskin2020curl}            & C      & DQN, SAC       & $o_t$         & $o_t$                 & \n  & \y  & \y  & \n  & \n  & J \\
CPC~\cite{oord2018representation}             & C      & A2C       & $o_t$         & $o_{t+k}$             & \n  & \n  & \n  & \n  & \n  & J \\
Bottleneck~\cite{yingjun2019learning} & C     & A2C       & $o_t$         & $o_{t+k}$             & \y  & \n  & \n  & \n  & \n  & J \\
DRIML~\cite{mazoure2020deep}    & C      & C51         & $o_t$         & $o_{t+k}$             & \y  & \n  & \n  & \n  & \n  & J \\
PI-SAC~\cite{lee2020predictive}   & C      & SAC       & $o_t$         & $o_{t+k}, r_{t+k}$    & \y  & \y  & \y  & \y  & \y  & J \\
ATC~\cite{stooke2020decoupling}      & C      & SAC, PPO         & $o_t$         & $o_{t+k}$             & \n  & \y  & \y  & \y  & \n  & P \\
PBL~\cite{guo2020bootstrap}      & B     & PopArt-IMPALA& $o_{1:t}$     & $o_{t+k}$             & \y & \n  & \n  & \y   & \n & J  \\
SPR~\cite{schwarzer2020data}      & B     & DQN         & $o_t$         & $o_{t+1:T}$                 & \y  & \y  & \y  & \y  & \n  & J \\
M-CURL~\cite{zhu2020masked} & C & DQN, SAC & $o_{1:t}$ & $o_t$ & \n & \y & \y & \n & \n & J\\
PlayVirtual~\cite{yu2021playvirtual} & B & DQN, SAC & $o_t$ & $o_{t+1:T}$ & \y & \y & \y & \y & \n & J\\
\bottomrule
\end{tabular}
\end{center}
\end{table}
\endgroup

\section{Additional information on experiment setup}
\label{sec:appendix-experiment}
\subsection{Environment setup}
\prg{DMC.} The training set for each DMC task consist of $250$ trajectories produced by an expert policy trained with RAD~\cite{laskin2020reinforcement}. For each of our methods we report the mean return for the final policy, which ranges between 0 and 1,000.

\prg{Procgen.} For each Procgen task, we reserve 100 environment seeds as ``training levels'', then use a separate 100 seeds as ``testing levels'' to evaluate generalization.
The training set for each task consists of 110K frames produced by applying RAD's trained agent to the training levels.
We report mean returns on the train and test levels separately in our results.

\prg{MAGICAL.} We give the IL and RepL algorithms access to five human demonstrations from the demo variant, then provide RepL with an additional 150,000 time steps of random rollouts (1,250 to 3,750 trajectories, depending on the environment).
This simulates situations where human expert demonstrations are expensive to collect, but robot exploration is relatively cheap.
For space reasons, we report mean scores averaged across all variants, which range between 0 and 1; full results are in \cref{app:magical-results}.

\subsection{RepL algorithms}
As is discussed in the main text, we evaluated 5 RepL algorithms. SimCLR and TemporalCPC are contrastive baselines: SimCLR must assign similar representations to augmented copies of the same observation, and different representations to augmented copies of different observations.
TemporalCPC must additionally account for dynamics by assigning similar representations to any pair of frames that are separated by a gap of $\Delta t$ time steps (we use $\Delta t=8$).
Like SimCLR, the VAE attempts to represent one frame at a time, without using temporal offsets. 
We also include methods that explicitly condition on or generate actions: dynamics predicts the next observation given the current observation and action, while inverse dynamics predicts which action was used to transition between two adjacent states.
Note that we do not use momentum, projection heads, or compression in our final experiments, since our preliminary experiments did not show a significant advantage to doing so.

\section{Related benchmarking work in RL and IL}
\label{sec:app-related-work}
\setlength{\tabcolsep}{5pt} 
\begin{table}[H]
\caption{Summary of previous works comparing different representation learning (repL) algorithms on imitation learning (IL) and reinforcement learning (RL). These works differ along a few axis on whether they provide a design breakdown of repL algorithms, their area of focus, the number of benchmarks covered, the number of algorithms experimented, whether they evaluate image-based environments, and whether they compare their results with an image augmentation baseline.
}
\label{table:related-work}
\begin{center}
\begin{tabular}{@{}lcccccc@{}}
\toprule
\bf{Research}                      & \bf{Design Breakdown} & \bf{Area} & \bf{\# of Benchmarks} & \bf{\# of Algo.} & \bf{Image env.} & \bf{Aug} \\
\midrule
ACL \cite{yang2021representation}  & \y                    & RL, IL    & 1                    & 10                & \n         & \n        \\
DrQ \cite{kostrikov2020image}      & \n                    & RL        & 1                    & 6                 & \y         & \y        \\
RAD \cite{laskin2020reinforcement} & \n                    & RL        & 3                    & 8                 & \y         & \y        \\
EIRLI (ours)                       & \y                    & IL        & 3                    & 7                 & \y         & \y       \\
\bottomrule
\end{tabular}
\end{center}
\end{table}

To the best of our knowledge, there are three existing work that survey RepL methods on imitation learning and reinforcement learning. We summarize their differences in Table \ref{table:related-work}. ACL \cite{yang2021representation} provided a design breakdown of different RepL algorithms too, and they found that many RepL algorithms perform poorly on imitation learning but can provide extra benefits for offline RL. DrQ \cite{kostrikov2020image} and RAD \cite{laskin2020reinforcement} both discussed the effect of image augmentation in reinforcement learning in great detail, and showed that with well-tuned image augmentations, a standard reinforcement learning framework can outperform many self-supervised learning methods on RL.

\section{Hyperparameter details} \label{appendix:hyperparams}

%


\begin{table}[H]
    \centering
    \begin{tabular}{@{}lcc@{}}
    \toprule
    \textbf{Hyperparameter}      & \textbf{Value}   & \textbf{Tuning range}        \\ \midrule
    \multicolumn{3}{c}{\textbf{All algorithms}}\\
    Optimizer & Adam & - \\
    LR & $10^{-4}$ & $10^{-6}$--$10^{-2}$\\
    Training batches & 5,000 & -\\
    Representation dim. & 128 & 64--256\\\midrule
    \multicolumn{3}{c}{\textbf{VAE, dyn., inv. dyn.}}\\
    Batch size & 64 &  - \\
    Augmentations & - & - \\\midrule
    \multicolumn{3}{c}{\textbf{VAE}}\\
    VAE $\beta$ & $10^{-6}$ & $10^{-7}$--$1.0$\\\midrule
    \multicolumn{3}{c}{\textbf{SimCLR, TCPC}}\\
    Batch size & 384 &  64--512 \\
    Augmentations & trans. rot., blur, col. jit.& -\\\midrule
    \multicolumn{3}{c}{\textbf{TCPC}}\\
    Temporal offset & 8 steps & - \\
    \bottomrule
    \end{tabular}
    \vspace{15pt}
    \caption{Hyperparameters for representation learning.
    Note that for joint training, BC and RepL use the same optimizer, and thus have the same learning rate.}
    \label{tab:repl-hps}
\end{table}

\begin{table}[H]
    \centering
    \begin{tabular}{@{}lcc@{}}
    \toprule
    \textbf{Hyperparameter}      & \textbf{Value}   & \textbf{Tuning range} \\ \midrule
    \multicolumn{3}{c}{\textbf{All benchmarks}}\\
    Optimizer & Adam & -\\
    LR & $10^{-4}$ & -\\
    Entropy coeff. & $10^{-3}$ & -\\
    $\ell_2$ reg. coeff. &  $10^{-5}$ & -\\
    Augmentations & trans., rot., blur, col. jit. & -\\
    \midrule
    \multicolumn{3}{c}{\textbf{All benchmarks, pretraining}}\\
    Batch size & 32 & -\\
    \midrule
    \multicolumn{3}{c}{\textbf{All benchmarks, joint training}}\\
    Batch size & 64 & -\\
    \midrule
    \multicolumn{3}{c}{\textbf{dm\_control and Procgen, pretraining and joint training}}\\
    Training batches & 1M & 1M--4M\\
    \midrule
    \multicolumn{3}{c}{\textbf{MAGICAL, pretraining}}\\
    Training batches & 20k &  5k--20k\\
    \midrule
    \multicolumn{3}{c}{\textbf{MAGICAL, joint training}}\\
    Training batches & 30k &  -\\
    \bottomrule
    \end{tabular}
    \vspace{15pt}
    \caption{
      Hyperparameters for behavioral cloning.
      Sections marked ``pretraining'' show the hyperparameters used for BC \textit{after} pretraining; sections marked ``joint training'' apply to BC during joint training.
      Representation learning hyperparameters, such as the batch size, are covered separately in \cref{tab:repl-hps}.}
    \label{tab:il-hps}
\end{table}

\begin{table}[t]
    \centering
    \begin{tabular}{lcccc}
        \toprule
        \textbf{Hyperparameter} & \textbf{DMC} & \textbf{Procgen} & \textbf{MAGICAL} & \textbf{Tuning range}\\\midrule
        Policy (PPO) & & & & \\
        \quad{}\# parallel envs & 32 & 32 & 32 & -\\
        \quad{}Time steps per round & 8 & 10 & 7 & 4--12\\
        \quad{}Epochs per round & 12 & 9 & 7 & 4--12 \\
        \quad{}Adam minibatch size & 48 & 48 & 48 & -\\
        \quad{}Initial Adam step size & $10^{-4}$ & $10^{-4}$ & $2.5 \times 10^{-4}$ & $5 \times (10^{-5}$--$10^{-4})$\\
        \quad{}Final Adam step size & \multicolumn{3}{c}{Linearly annealed to 0 over training} & -\\
        \quad{}Discount $\gamma$ & 0.99 & 0.6 & 0.99 & 0.6--1 \\
        \quad{}GAE $\lambda$ & 0.8 & 0.6 & 0.76 & 0.6--0.9\\
        \quad{}Entropy bonus & $10^{-8}$ & $5 \times 10^{-6}$ & $4.5 \times 10^{-8}$ & $10^{-10}$--$10^{-3}$\\
        \quad{}Advantage clip $\epsilon$ & 0.02 & 0.01 & 0.006 & $0.001$--$0.1$\\
        \quad{}Grad. clip $\ell_2$ norm & 1 & 1 & 1 & - \\
        \quad{}Augmentations & - & - & - & - \\
        Discriminator & & & & \\
        \quad{}Batch size & 48 & 48 & 48 & -\\
        \quad{}Adam step size & $10^{-3}$ & $2.5 \times 10^{-3}$ & $5.7 \times 10^{-4}$ & $5 \times (10^{-4}$--$10^{-3})$ \\
        \quad{}Disc. steps per round & 6 & 2 & 2 & 1--8\\
        \quad{}Augmentations & \makecell{Erase, blur, \\ noise, rot.} & \makecell{Col. jit., \\ flip LR, blur, \\ noise, rot., \\ trans.} & \makecell{Col. jit., \\ erase, flip LR, \\ blur, noise, \\ rot.} & \makecell{Col. jit., \\ erase, flip LR, \\ blur, noise, \\ rot., trans.}\\
        Misc. & &\\
        \quad{}Total env. steps of training & $5 \times 10^5$ & $5 \times 10^5$ & $5 \times 10^5$ & - \\
        \quad{}Reward norm. std. dev. & 0.01 & 0.01 & 0.01 & - \\
        \bottomrule
    \end{tabular}
    \vspace{2mm}
    \caption{
      Hyperparameters for GAIL experiments.
      We use the word ``round'' to describe the repeated sequence of data collection, followed by PPO updates on the collected data, followed by discriminator updates on both demonstrations and rollouts.
      Representation learning hyperparameters, such as the batch size, are covered separately in \cref{tab:repl-hps}.}
    \label{tab:hp-gail}
\end{table}

\prg{Environments and datasets}
For each dm\_control environment, we generated synthetic demonstration data using RAD with default algorithm hyperparameters~\cite{laskin2020reinforcement}.\footnote{\url{https://github.com/MishaLaskin/rad}}
Environment configurations (such as action repeat, frame stack, etc.) were the same for both RAD and our IL algorithms.
Specifically:
\begin{itemize}
    \item In cheetah-run, we used an action repeat of 4, resulting in a trajectory length of $1000/4=250$.
    Our demonstration dataset consisted of 250 trajectories (62,500 time steps) from the RAD demonstration agent, with a mean return of $\approx$827 (recall that return ranges between 0 and 1,000 for all \dmc{} environments).
    \item In finger-spin, we used an action repeat of 2, resulting in a trajectory length of $1000/2=500$.
    Our dataset again consisted of 250 trajectories (125,000 time steps) sampled from the RAD demonstration agent, with mean return of $\approx$963.
    \item In reacher-easy, we used an action repeat of 8, resulting in a trajectory length of $1000/8=125$.
    Our dataset of 250 trajectories (31,250 time steps) had mean return $\approx$977, and was again generated by RAD.
\end{itemize}
For all \dmc{} environments, we used a frame stack of 3.

As with \dmc{}, we generated expert demonstrations for Procgen using a policy trained with RAD.\footnote{\url{https://github.com/pokaxpoka/rad_procgen}}
We used the \texttt{easy} variants of all environments, with a frame stack of 3 and no action repeat.
We used a demonstration dataset of around 114,000 timesteps for each agent.
The mean trajectory lengths and returns are as follows:
\begin{itemize}
    \item For CoinRun, trajectories averaged 26 steps, and the demonstrator had an average return of 8.7.
    \item For Fruitbot, trajectories had an average length of 442, and the demonstrator attained a mean return of 29.75.
    \item For Jumper, trajectories had an average length of 76, and mean return of 8.7.
\end{itemize}

For each MAGICAL environment, we used a fixed subset of five demonstration trajectories (initially selected at random) from the human dataset provided with the benchmark~\cite{toyer2020magical}.
We used egocentric views with a frame stack of four and no action repeat.
Because there was no action repeat, trajectory lengths remained at the defaults for the benchmark suite: 40 for MoveToRegion, 80 for MoveToCorner, 120 for MatchRegions.
For each benchmark, we used between 25 and 28 demonstration trajectories, and the demonstrator attained the maximum return of 1.0 (on a 0.0--1.0 scale) in each trajectory.
In addition to demonstrations, our MAGICAL experiments also used random rollout datasets of 150,000 timesteps, all generated by uniformly sampling from the action set at each time step.
This equates to between 1,250 and 3,750 trajectories, depending on the horizon of the task.

\prg{RepL hyperparameters}
Representation learning hyperparameters are given in \cref{tab:repl-hps}.
Note that the contrastive algorithms have slightly different hyperparameters from the other RepL algorithms.
We found that a batch size close to 400 was important for contrastive algorithm performance; setting this value too low \textit{or} too high (e.g. 500+) decreased performance.
Predictive and reconstructive algorithms are less sensitive to batch size, so we used a computationally convenient batch size (64).

For the contrastive algorithms, we used a mixture of translation (trans.), rotation (rot.), Gaussian blur (blur), and color jitter (col. jit.) augmentations.
The translation augmentation translates the image by up to 5\% of image dimensions; the rotation augmentation rotates the image by up to 5$^{\circ{}}$; the Gaussian blur augmentation applies a Gaussian blur kernel with $\sigma=1$px; and the color jitter augmentation randomizes the hue by up to 0.15 radians.
We did not find the algorithms were highly sensitive to the choice of augmentations, but these augmentations did appear to perform fractionally better than the other choices that we considered during manual tuning.
For non-contrastive algorithms, we did not use augmentations.

For the VAE, we used a mean squared error loss for reconstruction, and down-weighted the KL prior term by a factor of $\beta$.
Specifically, our loss was
\[
    \mathcal L_{\text{VAE}} = \frac{1}{n} \sum_{i=1}^n (x_i - y_i)^2 - \beta \operatorname{KL}(e_\theta(\cdot \given x_i)\|\mathcal N(\cdot; 0, I))~,
\]
where $i$ indexes over the $n$ elements of the output image.
This is equivalent to using a traditional $\beta$-VAE with a fixed variance of $\sigma^2 = \frac{n}{2}$ for the Gaussian output distribution.

\prg{IL hyperparameters}
Imitation learning hyperparameters are given in \cref{tab:il-hps} for BC, and \cref{tab:hp-gail} for GAIL.
IL hyperparameters were the same for both control and RepL runs, except for the network initialization, where RepL runs initialized from the RepL-trained encoder, while control runs used a fresh He initialization~\cite{he2015delving}.
We found that \dmc{} and Procgen needed substantially more training than MAGICAL; we found that MAGICAL performance was plateauing even with an order of magnitude less training.

\prg{Compute information}
Running one experiment (a single seed of combined RepL and IL) on \dmc{} and Procgen for 1M batch updates takes about 40 hours on one NVIDIA 1080Ti, and running one for MAGICAL between 30 minutes (for 20,000-batch control without augmentations) and 10 hours (for a contrastive method using joint training for 30,000 batches). Generating the results in this paper takes around 6,600 GPU hours for \dmc{}, 8,800 GPU hours for Procgen, and 26--525 hours for MAGICAL (assuming 4 seeds per GPU, and 30 minutes to 10 hours per seed).

\section{Limitations, social impacts, and benchmark license}

\paragraph{Limitations} The main limitation of our findings is that we only investigate policy learning with BC.
Our findings therefore may not generalise to IL algorithms that learn more than just a policy.
This includes IRL algorithms, which typically learn both a reward function and a policy, as well as IL algorithms like SQIL~\cite{reddy2019sqil} that learn a Q function rather than directly learning a policy.

\paragraph{Social impacts} We do not foresee any negative near-term social impact from our work.

\paragraph{License} We release our codebase and associated data under the MIT license.

\section{Implementation of components in the codebase}
In Section \ref{sec:repl-analysis} we analyzed several design axes and their components. We elaborate in this section our current implementation status of these components in the codebase.

\prg{Target selection.} Different versions of this design choice are implemented by inheriting from the \texttt{TargetPairConstructor} class within the codebase. Currently implemented are \texttt{IdentityPairConstructor}, in which the context and target are identical, and \texttt{TemporalOffsetPairConstructor}, which can be given a desired temporal offset, and configured to optionally return the action as extra context. 

\prg{Loss type.} Different versions of loss functions are implemented by inheriting from \texttt{RepresentationLoss}. We have already implemented a wide variety of losses, including VAE, mean squared error, negative log likelihood, CEB, and several contrastive losses.

\prg{Augmentation.} This design choice is implemented in subclasses of \texttt{Augmenter}. We rely on a standard library to implement the augmentations. Each subclass augments a different set of inputs: both the context and target, only the context, or neither the context nor target.

\prg{Encoder.} Different versions of encoder are implemented by inheriting from the \texttt{Encoder} class within the codebase. We have implemented encoders that work on individual images, as well as a \texttt{RecurrentEncoder}.

\prg{Decoder.} Different versions of decoder are implemented by inheriting from the \texttt{LossDecoder} class within the codebase. Currently implemented decoders support image reconstruction, action conditioning, and projection heads.

\section{Complete MAGICAL results}\label{app:magical-results}
\begin{table}[H]
    \centering
    \begin{footnotesize}
        \begin{adjustbox}{center}
            \begin{tabular}{@{}lccccccc@{}}
                \toprule
                \textbf{Task} & \textbf{Dynamics} & \textbf{InvDyn} & \textbf{SimCLR} & \textbf{TemporalCPC} & \textbf{VAE} & \textbf{BC, augs} & \textbf{BC, no augs} \\
                \midrule
                MatchRegions-Demo & 0.71$\mpm$0.06 & 0.72$\mpm$0.07 & 0.72$\mpm$0.06 & 0.68$\mpm$0.03 & 0.70$\mpm$0.05 & 0.76$\mpm$0.05 & 0.63$\mpm$0.15 \\
                \ \ \ \ -TestDynamics & 0.55$\mpm$0.09 & 0.56$\mpm$0.07 & 0.56$\mpm$0.07 & 0.52$\mpm$0.01 & 0.56$\mpm$0.08 & 0.66$\mpm$0.05 & 0.38$\mpm$0.12 \\
                \ \ \ \ -TestColour & \cellcolor[HTML]{fff7df}0.71$\mpm$0.06* & \cellcolor[HTML]{fff7df}0.70$\mpm$0.08* & \cellcolor[HTML]{fff7df}0.67$\mpm$0.04* & \cellcolor[HTML]{fff7df}0.66$\mpm$0.04* & \cellcolor[HTML]{fff7df}0.67$\mpm$0.06* & 0.53$\mpm$0.04 & 0.30$\mpm$0.08 \\
                \ \ \ \ -TestShape & 0.61$\mpm$0.06 & 0.64$\mpm$0.09 & 0.62$\mpm$0.06 & 0.61$\mpm$0.04 & 0.60$\mpm$0.05 & 0.70$\mpm$0.03 & 0.43$\mpm$0.13 \\
                \ \ \ \ -TestJitter & \cellcolor[HTML]{fff7df}0.70$\mpm$0.06 & 0.65$\mpm$0.06 & 0.66$\mpm$0.07 & 0.65$\mpm$0.04 & \cellcolor[HTML]{fff7df}0.68$\mpm$0.03 & 0.68$\mpm$0.05 & 0.41$\mpm$0.13 \\
                \ \ \ \ -TestLayout & 0.04$\mpm$0.01 & 0.04$\mpm$0.01 & 0.04$\mpm$0.01 & \cellcolor[HTML]{fff7df}0.05$\mpm$0.01 & 0.04$\mpm$0.01 & 0.05$\mpm$0.01 & 0.05$\mpm$0.03 \\
                \ \ \ \ -TestCountPlus & 0.04$\mpm$0.02 & 0.04$\mpm$0.01 & \cellcolor[HTML]{fff7df}0.06$\mpm$0.03 & 0.05$\mpm$0.01 & 0.04$\mpm$0.02 & 0.05$\mpm$0.02 & 0.04$\mpm$0.02 \\
                \ \ \ \ -TestAll & 0.04$\mpm$0.02 & \cellcolor[HTML]{fff7df}0.05$\mpm$0.02 & \cellcolor[HTML]{fff7df}0.05$\mpm$0.03 & \cellcolor[HTML]{fff7df}0.06$\mpm$0.01* & 0.04$\mpm$0.02 & 0.05$\mpm$0.01 & 0.03$\mpm$0.02 \\
                \ \ \ \ Average & 0.42$\mpm$0.04 & 0.42$\mpm$0.04 & 0.42$\mpm$0.03 & 0.41$\mpm$0.01 & 0.42$\mpm$0.03 & 0.43$\mpm$0.02 & 0.28$\mpm$0.08 \\
                \midrule
                MoveToCorner-Demo & \cellcolor[HTML]{fff7df}0.94$\mpm$0.08 & \cellcolor[HTML]{fff7df}0.92$\mpm$0.07 & \cellcolor[HTML]{fff7df}0.88$\mpm$0.09 & \cellcolor[HTML]{fff7df}0.86$\mpm$0.10 & \cellcolor[HTML]{fff7df}0.89$\mpm$0.08 & 0.86$\mpm$0.08 & \cellcolor[HTML]{fff7df}0.99$\mpm$0.01* \\
                \ \ \ \ -TestDynamics & \cellcolor[HTML]{fff7df}0.83$\mpm$0.05 & \cellcolor[HTML]{fff7df}0.87$\mpm$0.05* & \cellcolor[HTML]{fff7df}0.85$\mpm$0.04* & \cellcolor[HTML]{fff7df}0.80$\mpm$0.06 & 0.75$\mpm$0.04 & 0.76$\mpm$0.08 & 0.68$\mpm$0.04 \\
                \ \ \ \ -TestColour & \cellcolor[HTML]{fff7df}0.90$\mpm$0.10 & \cellcolor[HTML]{fff7df}0.87$\mpm$0.06 & \cellcolor[HTML]{fff7df}0.86$\mpm$0.13 & \cellcolor[HTML]{fff7df}0.86$\mpm$0.06 & \cellcolor[HTML]{fff7df}0.87$\mpm$0.07 & 0.75$\mpm$0.16 & \cellcolor[HTML]{fff7df}0.85$\mpm$0.07 \\
                \ \ \ \ -TestShape & \cellcolor[HTML]{fff7df}0.90$\mpm$0.06 & \cellcolor[HTML]{fff7df}0.88$\mpm$0.07 & \cellcolor[HTML]{fff7df}0.91$\mpm$0.05* & \cellcolor[HTML]{fff7df}0.87$\mpm$0.02 & \cellcolor[HTML]{fff7df}0.84$\mpm$0.09 & 0.84$\mpm$0.06 & 0.71$\mpm$0.10 \\
                \ \ \ \ -TestJitter & \cellcolor[HTML]{fff7df}0.81$\mpm$0.09 & \cellcolor[HTML]{fff7df}0.80$\mpm$0.04 & \cellcolor[HTML]{fff7df}0.84$\mpm$0.04* & \cellcolor[HTML]{fff7df}0.79$\mpm$0.04 & 0.77$\mpm$0.04 & 0.78$\mpm$0.00 & 0.58$\mpm$0.08 \\
                \ \ \ \ -TestAll & 0.66$\mpm$0.13 & 0.64$\mpm$0.09 & 0.65$\mpm$0.07 & 0.62$\mpm$0.06 & 0.57$\mpm$0.12 & 0.67$\mpm$0.09 & 0.51$\mpm$0.13 \\
                \ \ \ \ Average & \cellcolor[HTML]{fff7df}0.84$\mpm$0.07 & \cellcolor[HTML]{fff7df}0.83$\mpm$0.04 & \cellcolor[HTML]{fff7df}0.83$\mpm$0.04* & \cellcolor[HTML]{fff7df}0.80$\mpm$0.02 & \cellcolor[HTML]{fff7df}0.78$\mpm$0.06 & 0.78$\mpm$0.05 & 0.72$\mpm$0.04 \\
                \midrule
                MoveToRegion-Demo & 0.99$\mpm$0.01 & 0.99$\mpm$0.01 & 1.00$\mpm$0.01 & 0.98$\mpm$0.02 & 0.99$\mpm$0.02 & 1.00$\mpm$0.00 & \cellcolor[HTML]{fff7df}1.00$\mpm$0.00 \\
                \ \ \ \ -TestDynamics & 0.99$\mpm$0.01 & 1.00$\mpm$0.01 & 0.99$\mpm$0.01 & 0.99$\mpm$0.01 & 0.99$\mpm$0.01 & 1.00$\mpm$0.01 & \cellcolor[HTML]{fff7df}1.00$\mpm$0.00 \\
                \ \ \ \ -TestColour & \cellcolor[HTML]{fff7df}0.99$\mpm$0.02* & \cellcolor[HTML]{fff7df}0.99$\mpm$0.01* & \cellcolor[HTML]{fff7df}1.00$\mpm$0.00* & \cellcolor[HTML]{fff7df}0.99$\mpm$0.01* & \cellcolor[HTML]{fff7df}0.97$\mpm$0.04* & 0.54$\mpm$0.08 & \cellcolor[HTML]{fff7df}0.84$\mpm$0.13* \\
                \ \ \ \ -TestJitter & 0.99$\mpm$0.01 & 1.00$\mpm$0.00 & \cellcolor[HTML]{fff7df}1.00$\mpm$0.00 & \cellcolor[HTML]{fff7df}1.00$\mpm$0.00 & 1.00$\mpm$0.01 & 1.00$\mpm$0.00 & 0.99$\mpm$0.01 \\
                \ \ \ \ -TestLayout & 0.52$\mpm$0.05 & 0.51$\mpm$0.05 & 0.45$\mpm$0.03 & 0.44$\mpm$0.07 & 0.46$\mpm$0.11 & 0.64$\mpm$0.03 & 0.64$\mpm$0.11 \\
                \ \ \ \ -TestAll & \cellcolor[HTML]{fff7df}0.45$\mpm$0.06* & \cellcolor[HTML]{fff7df}0.50$\mpm$0.09* & \cellcolor[HTML]{fff7df}0.46$\mpm$0.06* & \cellcolor[HTML]{fff7df}0.42$\mpm$0.08* & \cellcolor[HTML]{fff7df}0.47$\mpm$0.12* & 0.28$\mpm$0.06 & \cellcolor[HTML]{fff7df}0.38$\mpm$0.08* \\
                \ \ \ \ Average & \cellcolor[HTML]{fff7df}0.82$\mpm$0.02* & \cellcolor[HTML]{fff7df}0.83$\mpm$0.02* & \cellcolor[HTML]{fff7df}0.82$\mpm$0.01* & \cellcolor[HTML]{fff7df}0.81$\mpm$0.01* & \cellcolor[HTML]{fff7df}0.81$\mpm$0.05* & 0.74$\mpm$0.02 & \cellcolor[HTML]{fff7df}0.81$\mpm$0.04* \\
                \bottomrule
            \end{tabular}
        \end{adjustbox}
    \end{footnotesize}
    \caption{
      Complete results for all variants of the evaluated MAGICAL tasks, with BC plus RepL pretraining.
      Refer to \cref{tab:magical-full-results-jt} for joint training.
    }
    \label{tab:magical-full-results-pretrain}
\end{table}%
\begin{table}[H]
    \centering
    \begin{footnotesize}
        \begin{adjustbox}{center}
            \begin{tabular}{@{}lccccccc@{}}
                \toprule
                \textbf{Task} & \textbf{Dynamics} & \textbf{InvDyn} & \textbf{SimCLR} & \textbf{TemporalCPC} & \textbf{VAE} & \textbf{BC, augs} & \textbf{BC, no augs} \\
                \midrule
                MatchRegions-Demo & \cellcolor[HTML]{fff7df}0.78$\mpm$0.03 & 0.51$\mpm$0.16 & 0.73$\mpm$0.06 & 0.01$\mpm$0.01 & 0.72$\mpm$0.09 & 0.78$\mpm$0.06 & 0.70$\mpm$0.04 \\
                \ \ \ \ -TestDynamics & 0.62$\mpm$0.04 & 0.39$\mpm$0.14 & 0.59$\mpm$0.04 & 0.01$\mpm$0.01 & 0.58$\mpm$0.05 & 0.63$\mpm$0.05 & 0.40$\mpm$0.03 \\
                \ \ \ \ -TestColour & \cellcolor[HTML]{fff7df}0.60$\mpm$0.03* & 0.17$\mpm$0.07 & \cellcolor[HTML]{fff7df}0.55$\mpm$0.06* & 0.01$\mpm$0.01 & \cellcolor[HTML]{fff7df}0.48$\mpm$0.04 & 0.47$\mpm$0.08 & 0.35$\mpm$0.10 \\
                \ \ \ \ -TestShape & \cellcolor[HTML]{fff7df}0.71$\mpm$0.04* & 0.36$\mpm$0.14 & \cellcolor[HTML]{fff7df}0.67$\mpm$0.05 & 0.01$\mpm$0.01 & \cellcolor[HTML]{fff7df}0.68$\mpm$0.05 & 0.66$\mpm$0.04 & 0.48$\mpm$0.02 \\
                \ \ \ \ -TestJitter & \cellcolor[HTML]{fff7df}0.69$\mpm$0.05 & 0.36$\mpm$0.14 & 0.66$\mpm$0.04 & 0.01$\mpm$0.01 & 0.65$\mpm$0.04 & 0.66$\mpm$0.06 & 0.40$\mpm$0.10 \\
                \ \ \ \ -TestLayout & 0.05$\mpm$0.01 & 0.04$\mpm$0.01 & 0.04$\mpm$0.01 & 0.01$\mpm$0.01 & 0.05$\mpm$0.02 & 0.07$\mpm$0.01 & 0.03$\mpm$0.01 \\
                \ \ \ \ -TestCountPlus & 0.04$\mpm$0.01 & 0.02$\mpm$0.01 & 0.04$\mpm$0.01 & 0.01$\mpm$0.01 & 0.05$\mpm$0.03 & 0.07$\mpm$0.02 & 0.04$\mpm$0.01 \\
                \ \ \ \ -TestAll & 0.04$\mpm$0.01 & 0.02$\mpm$0.01 & 0.03$\mpm$0.01 & 0.01$\mpm$0.01 & 0.04$\mpm$0.01 & 0.07$\mpm$0.02 & 0.04$\mpm$0.03 \\
                \ \ \ \ Average & \cellcolor[HTML]{fff7df}0.44$\mpm$0.02 & 0.23$\mpm$0.08 & 0.41$\mpm$0.02 & 0.01$\mpm$0.01 & 0.41$\mpm$0.03 & 0.43$\mpm$0.03 & 0.31$\mpm$0.02 \\
                \midrule
                MoveToCorner-Demo & 0.86$\mpm$0.04 & 0.37$\mpm$0.25 & 0.90$\mpm$0.12 & 0.03$\mpm$0.03 & 0.94$\mpm$0.05 & 0.95$\mpm$0.04 & \cellcolor[HTML]{fff7df}0.99$\mpm$0.00* \\
                \ \ \ \ -TestDynamics & 0.82$\mpm$0.07 & 0.35$\mpm$0.27 & \cellcolor[HTML]{fff7df}0.86$\mpm$0.05 & 0.03$\mpm$0.03 & \cellcolor[HTML]{fff7df}0.87$\mpm$0.04 & 0.82$\mpm$0.06 & 0.75$\mpm$0.09 \\
                \ \ \ \ -TestColour & \cellcolor[HTML]{fff7df}0.84$\mpm$0.08 & 0.25$\mpm$0.29 & 0.61$\mpm$0.17 & 0.01$\mpm$0.01 & 0.73$\mpm$0.15 & 0.77$\mpm$0.15 & 0.76$\mpm$0.17 \\
                \ \ \ \ -TestShape & 0.84$\mpm$0.12 & 0.44$\mpm$0.27 & 0.88$\mpm$0.09 & 0.02$\mpm$0.03 & \cellcolor[HTML]{fff7df}0.93$\mpm$0.05 & 0.89$\mpm$0.04 & 0.76$\mpm$0.09 \\
                \ \ \ \ -TestJitter & 0.76$\mpm$0.06 & 0.25$\mpm$0.18 & 0.78$\mpm$0.06 & 0.02$\mpm$0.03 & \cellcolor[HTML]{fff7df}0.83$\mpm$0.07 & 0.81$\mpm$0.08 & 0.55$\mpm$0.12 \\
                \ \ \ \ -TestAll & \cellcolor[HTML]{fff7df}0.57$\mpm$0.11 & 0.13$\mpm$0.15 & 0.53$\mpm$0.10 & 0.01$\mpm$0.01 & \cellcolor[HTML]{fff7df}0.63$\mpm$0.09 & 0.57$\mpm$0.10 & 0.39$\mpm$0.21 \\
                \ \ \ \ Average & 0.78$\mpm$0.07 & 0.30$\mpm$0.22 & 0.76$\mpm$0.05 & 0.02$\mpm$0.02 & \cellcolor[HTML]{fff7df}0.82$\mpm$0.06 & 0.80$\mpm$0.05 & 0.70$\mpm$0.09 \\
                \midrule
                MoveToRegion-Demo & 1.00$\mpm$0.00 & 0.60$\mpm$0.34 & 1.00$\mpm$0.00 & 0.81$\mpm$0.14 & 1.00$\mpm$0.00 & 1.00$\mpm$0.00 & 1.00$\mpm$0.00 \\
                \ \ \ \ -TestDynamics & \cellcolor[HTML]{fff7df}1.00$\mpm$0.00 & 0.55$\mpm$0.37 & \cellcolor[HTML]{fff7df}1.00$\mpm$0.00 & 0.81$\mpm$0.14 & \cellcolor[HTML]{fff7df}1.00$\mpm$0.00 & 1.00$\mpm$0.00 & 0.99$\mpm$0.01 \\
                \ \ \ \ -TestColour & 0.64$\mpm$0.15 & 0.22$\mpm$0.15 & \cellcolor[HTML]{fff7df}0.93$\mpm$0.07* & 0.25$\mpm$0.06 & \cellcolor[HTML]{fff7df}0.69$\mpm$0.12 & 0.65$\mpm$0.13 & \cellcolor[HTML]{fff7df}0.77$\mpm$0.12 \\
                \ \ \ \ -TestJitter & 1.00$\mpm$0.00 & 0.49$\mpm$0.40 & 1.00$\mpm$0.00 & 0.78$\mpm$0.08 & 1.00$\mpm$0.00 & 1.00$\mpm$0.00 & 0.98$\mpm$0.01 \\
                \ \ \ \ -TestLayout & \cellcolor[HTML]{fff7df}0.64$\mpm$0.05 & 0.17$\mpm$0.16 & 0.30$\mpm$0.06 & 0.10$\mpm$0.05 & \cellcolor[HTML]{fff7df}0.68$\mpm$0.06* & 0.61$\mpm$0.04 & \cellcolor[HTML]{fff7df}0.63$\mpm$0.06 \\
                \ \ \ \ -TestAll & \cellcolor[HTML]{fff7df}0.29$\mpm$0.06 & 0.10$\mpm$0.08 & 0.23$\mpm$0.04 & 0.06$\mpm$0.03 & \cellcolor[HTML]{fff7df}0.28$\mpm$0.03* & 0.23$\mpm$0.03 & \cellcolor[HTML]{fff7df}0.31$\mpm$0.08 \\
                \ \ \ \ Average & \cellcolor[HTML]{fff7df}0.76$\mpm$0.02 & 0.35$\mpm$0.24 & 0.74$\mpm$0.01 & 0.47$\mpm$0.07 & \cellcolor[HTML]{fff7df}0.77$\mpm$0.02 & 0.75$\mpm$0.02 & \cellcolor[HTML]{fff7df}0.78$\mpm$0.04 \\
                \bottomrule
            \end{tabular}
        \end{adjustbox}
    \end{footnotesize}
    \caption{
      Complete results for all variants of the evaluated MAGICAL tasks, with BC plus RepL joint training.
      Refer to \cref{tab:magical-full-results-pretrain} for BC plus RepL pretraining.
    }
    \label{tab:magical-full-results-jt}
\end{table}
\begin{table}[H]
    \centering
    \begin{footnotesize}
        \begin{adjustbox}{center}
            \begin{tabular}{@{}lccccccc@{}}
                \toprule
                \textbf{Task} & \textbf{Dynamics} & \textbf{InvDyn} & \textbf{SimCLR} & \textbf{TemporalCPC} & \textbf{VAE} & \textbf{GAIL augs} & \textbf{GAIL no augs} \\
                \midrule
                MatchRegions-Demo & 0.76$\mpm$0.19 & 0.59$\mpm$0.24 & \cellcolor[HTML]{fff7df}0.83$\mpm$0.09 & 0.68$\mpm$0.22 & 0.51$\mpm$0.28 & 0.81$\mpm$0.10 & 0.36$\mpm$0.22 \\
                \ \ \ \ -TestDynamics & 0.74$\mpm$0.17 & 0.55$\mpm$0.22 & \cellcolor[HTML]{fff7df}0.79$\mpm$0.11 & 0.64$\mpm$0.25 & 0.51$\mpm$0.27 & 0.77$\mpm$0.13 & 0.37$\mpm$0.22 \\
                \ \ \ \ -TestColour & 0.21$\mpm$0.04 & 0.18$\mpm$0.06 & \cellcolor[HTML]{fff7df}0.25$\mpm$0.04 & 0.21$\mpm$0.07 & 0.14$\mpm$0.08 & 0.24$\mpm$0.03 & 0.08$\mpm$0.05 \\
                \ \ \ \ -TestShape & 0.72$\mpm$0.20 & 0.56$\mpm$0.22 & \cellcolor[HTML]{fff7df}0.81$\mpm$0.10 & 0.61$\mpm$0.25 & 0.49$\mpm$0.26 & 0.76$\mpm$0.10 & 0.35$\mpm$0.20 \\
                \ \ \ \ -TestJitter & 0.73$\mpm$0.17 & 0.57$\mpm$0.20 & \cellcolor[HTML]{fff7df}0.83$\mpm$0.10 & 0.67$\mpm$0.22 & 0.53$\mpm$0.28 & 0.79$\mpm$0.13 & 0.38$\mpm$0.22 \\
                \ \ \ \ -TestLayout & 0.11$\mpm$0.03 & 0.13$\mpm$0.04 & 0.11$\mpm$0.03 & 0.15$\mpm$0.06 & 0.13$\mpm$0.06 & 0.17$\mpm$0.04 & 0.12$\mpm$0.06 \\
                \ \ \ \ -TestCountPlus & 0.06$\mpm$0.03 & 0.07$\mpm$0.03 & \cellcolor[HTML]{fff7df}0.09$\mpm$0.05 & \cellcolor[HTML]{fff7df}0.08$\mpm$0.04 & 0.05$\mpm$0.02 & 0.08$\mpm$0.04 & 0.05$\mpm$0.02 \\
                \ \ \ \ -TestAll & 0.05$\mpm$0.01 & 0.06$\mpm$0.03 & 0.08$\mpm$0.02 & 0.08$\mpm$0.02 & 0.06$\mpm$0.02 & 0.09$\mpm$0.04 & 0.06$\mpm$0.03 \\
                \ \ \ \ Average & 0.42$\mpm$0.10 & 0.34$\mpm$0.12 & \cellcolor[HTML]{fff7df}0.47$\mpm$0.04 & 0.39$\mpm$0.12 & 0.30$\mpm$0.15 & 0.46$\mpm$0.06 & 0.22$\mpm$0.12 \\
                \midrule
                MoveToCorner-Demo & \cellcolor[HTML]{fff7df}0.67$\mpm$0.09 & 0.60$\mpm$0.09 & \cellcolor[HTML]{fff7df}0.67$\mpm$0.14 & \cellcolor[HTML]{fff7df}0.72$\mpm$0.15 & \cellcolor[HTML]{fff7df}0.78$\mpm$0.07* & 0.63$\mpm$0.13 & 0.62$\mpm$0.14 \\
                \ \ \ \ -TestDynamics & 0.59$\mpm$0.10 & 0.58$\mpm$0.10 & \cellcolor[HTML]{fff7df}0.66$\mpm$0.09 & \cellcolor[HTML]{fff7df}0.69$\mpm$0.11 & \cellcolor[HTML]{fff7df}0.76$\mpm$0.09* & 0.65$\mpm$0.12 & 0.64$\mpm$0.16 \\
                \ \ \ \ -TestColour & 0.35$\mpm$0.08 & 0.32$\mpm$0.16 & \cellcolor[HTML]{fff7df}0.39$\mpm$0.09 & \cellcolor[HTML]{fff7df}0.45$\mpm$0.21 & \cellcolor[HTML]{fff7df}0.51$\mpm$0.19 & 0.37$\mpm$0.10 & \cellcolor[HTML]{fff7df}0.57$\mpm$0.16* \\
                \ \ \ \ -TestShape & 0.43$\mpm$0.10 & 0.43$\mpm$0.16 & \cellcolor[HTML]{fff7df}0.51$\mpm$0.09 & \cellcolor[HTML]{fff7df}0.52$\mpm$0.16 & \cellcolor[HTML]{fff7df}0.54$\mpm$0.14 & 0.49$\mpm$0.12 & 0.47$\mpm$0.15 \\
                \ \ \ \ -TestJitter & 0.61$\mpm$0.10 & 0.54$\mpm$0.09 & \cellcolor[HTML]{fff7df}0.62$\mpm$0.12 & \cellcolor[HTML]{fff7df}0.63$\mpm$0.16 & \cellcolor[HTML]{fff7df}0.74$\mpm$0.08* & 0.62$\mpm$0.11 & 0.61$\mpm$0.16 \\
                \ \ \ \ -TestAll & 0.19$\mpm$0.09 & \cellcolor[HTML]{fff7df}0.21$\mpm$0.20 & \cellcolor[HTML]{fff7df}0.25$\mpm$0.14 & \cellcolor[HTML]{fff7df}0.31$\mpm$0.19 & \cellcolor[HTML]{fff7df}0.38$\mpm$0.18* & 0.20$\mpm$0.13 & \cellcolor[HTML]{fff7df}0.39$\mpm$0.16* \\
                \ \ \ \ Average & 0.48$\mpm$0.09 & 0.45$\mpm$0.10 & \cellcolor[HTML]{fff7df}0.52$\mpm$0.07 & \cellcolor[HTML]{fff7df}0.55$\mpm$0.15 & \cellcolor[HTML]{fff7df}0.62$\mpm$0.11* & 0.49$\mpm$0.08 & \cellcolor[HTML]{fff7df}0.55$\mpm$0.14 \\
                \midrule
                MoveToRegion-Demo & \cellcolor[HTML]{fff7df}0.96$\mpm$0.04 & 0.95$\mpm$0.04 & 0.95$\mpm$0.07 & \cellcolor[HTML]{fff7df}0.97$\mpm$0.03 & 0.96$\mpm$0.04 & 0.96$\mpm$0.08 & 0.89$\mpm$0.09 \\
                \ \ \ \ -TestDynamics & 0.94$\mpm$0.05 & 0.94$\mpm$0.05 & 0.94$\mpm$0.07 & \cellcolor[HTML]{fff7df}0.97$\mpm$0.03 & \cellcolor[HTML]{fff7df}0.95$\mpm$0.04 & 0.95$\mpm$0.06 & 0.82$\mpm$0.20 \\
                \ \ \ \ -TestColour & 0.65$\mpm$0.16 & \cellcolor[HTML]{fff7df}0.68$\mpm$0.10 & \cellcolor[HTML]{fff7df}0.69$\mpm$0.23 & \cellcolor[HTML]{fff7df}0.68$\mpm$0.12 & \cellcolor[HTML]{fff7df}0.68$\mpm$0.21 & 0.67$\mpm$0.19 & 0.57$\mpm$0.21 \\
                \ \ \ \ -TestJitter & 0.92$\mpm$0.06 & 0.93$\mpm$0.05 & \cellcolor[HTML]{fff7df}0.96$\mpm$0.08 & \cellcolor[HTML]{fff7df}0.97$\mpm$0.04 & \cellcolor[HTML]{fff7df}0.97$\mpm$0.03 & 0.94$\mpm$0.08 & 0.69$\mpm$0.25 \\
                \ \ \ \ -TestLayout & 0.59$\mpm$0.09 & 0.61$\mpm$0.10 & 0.59$\mpm$0.09 & \cellcolor[HTML]{fff7df}0.66$\mpm$0.06 & 0.64$\mpm$0.14 & 0.65$\mpm$0.16 & 0.39$\mpm$0.18 \\
                \ \ \ \ -TestAll & 0.28$\mpm$0.09 & \cellcolor[HTML]{fff7df}0.36$\mpm$0.09 & 0.28$\mpm$0.08 & 0.29$\mpm$0.06 & 0.32$\mpm$0.08 & 0.35$\mpm$0.10 & 0.24$\mpm$0.09 \\
                \ \ \ \ Average & 0.72$\mpm$0.07 & 0.74$\mpm$0.04 & 0.74$\mpm$0.06 & \cellcolor[HTML]{fff7df}0.76$\mpm$0.03 & 0.75$\mpm$0.07 & 0.75$\mpm$0.09 & 0.60$\mpm$0.14 \\
                \bottomrule
            \end{tabular}
        \end{adjustbox}
    \end{footnotesize}
    \caption{
      Complete results for all variants of the evaluated MAGICAL tasks, with GAIL plus RepL pretraining.
    }
    \label{tab:magical-full-results-pretrain-gail}
\end{table}%

\section{Loss curves}\label{app:loss-curves}
\subsection{Pretrain}
\begin{figure}[H]
    \begin{subfigure}{0.33\textwidth}
        \centering
        \includegraphics[width=\textwidth]{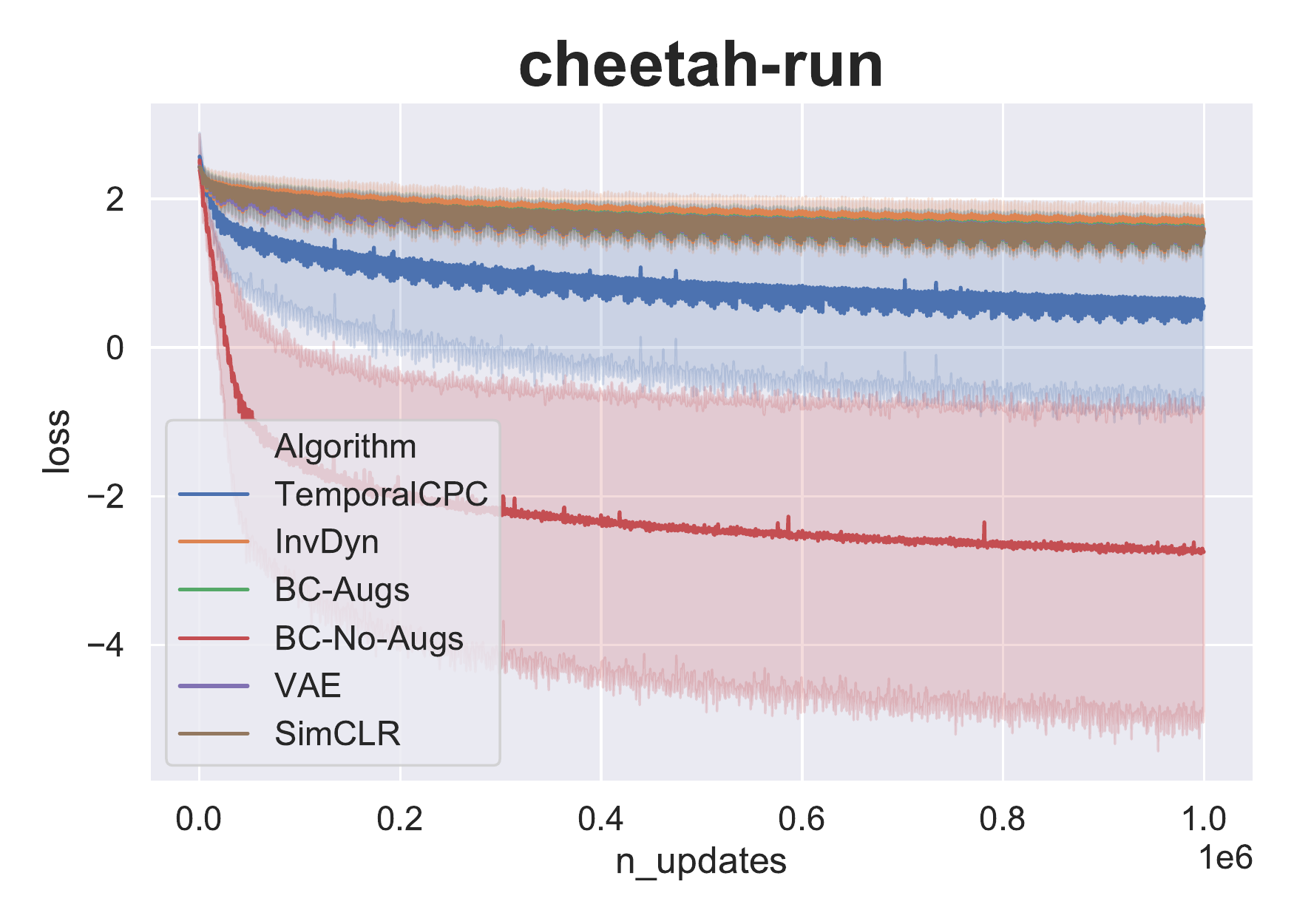}
    \end{subfigure}
    \begin{subfigure}{0.33\textwidth}
        \centering
        \includegraphics[width=\textwidth]{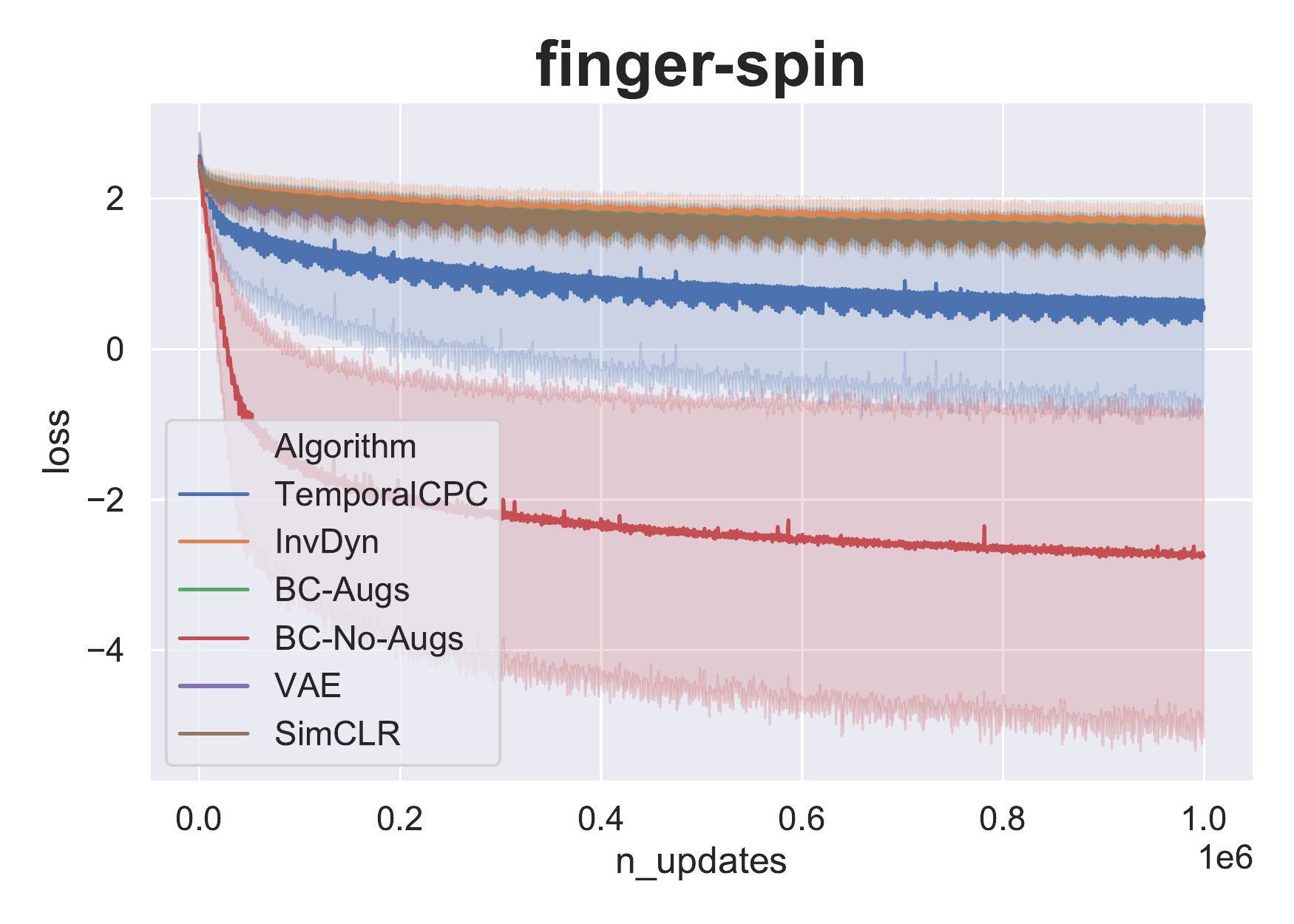}
    \end{subfigure}
    \begin{subfigure}{0.33\textwidth}
        \centering
        \includegraphics[width=\textwidth]{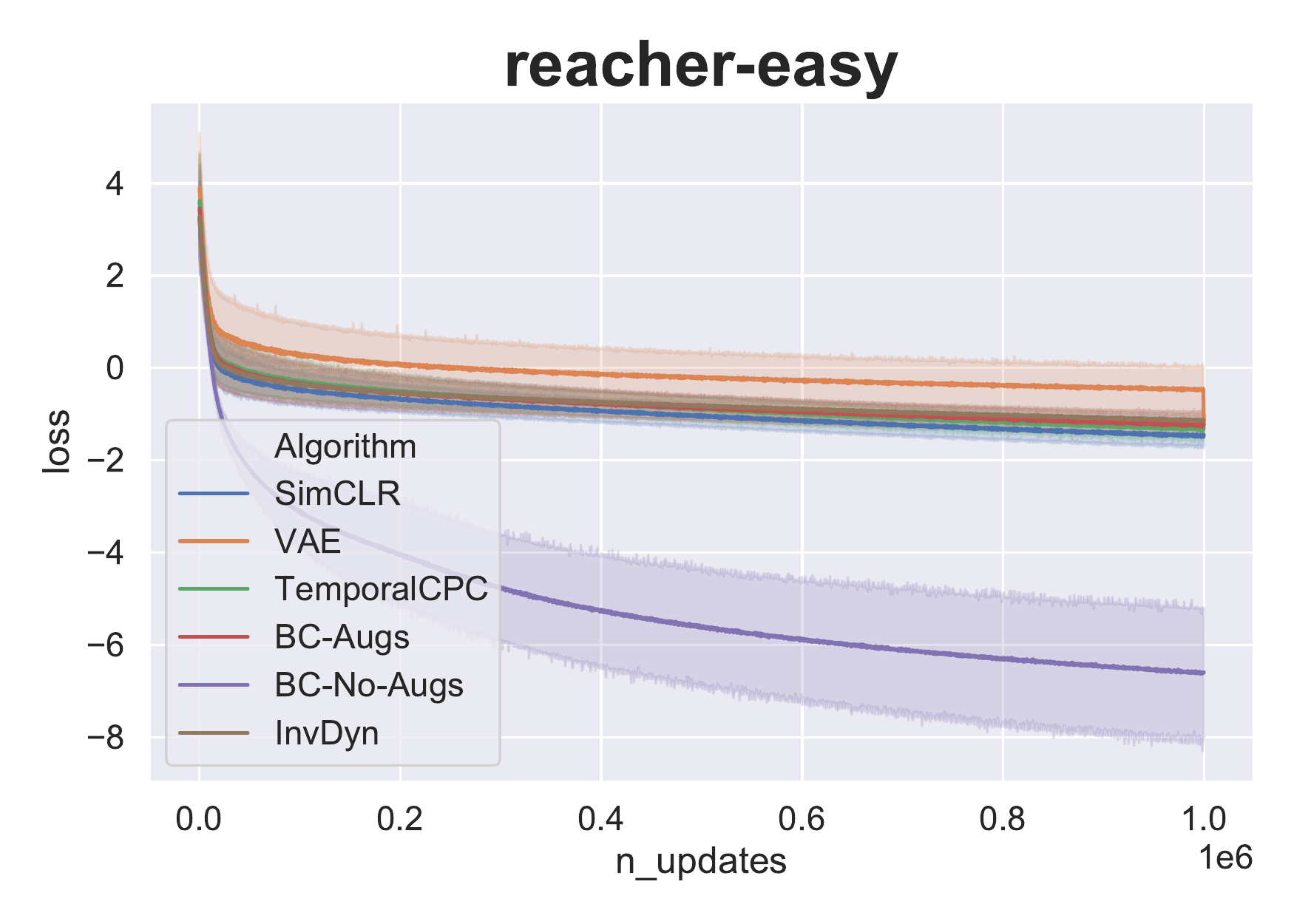}
    \end{subfigure}
\end{figure}
\begin{figure}[H]
    \begin{subfigure}{0.33\textwidth}
        \centering
        \includegraphics[width=\textwidth]{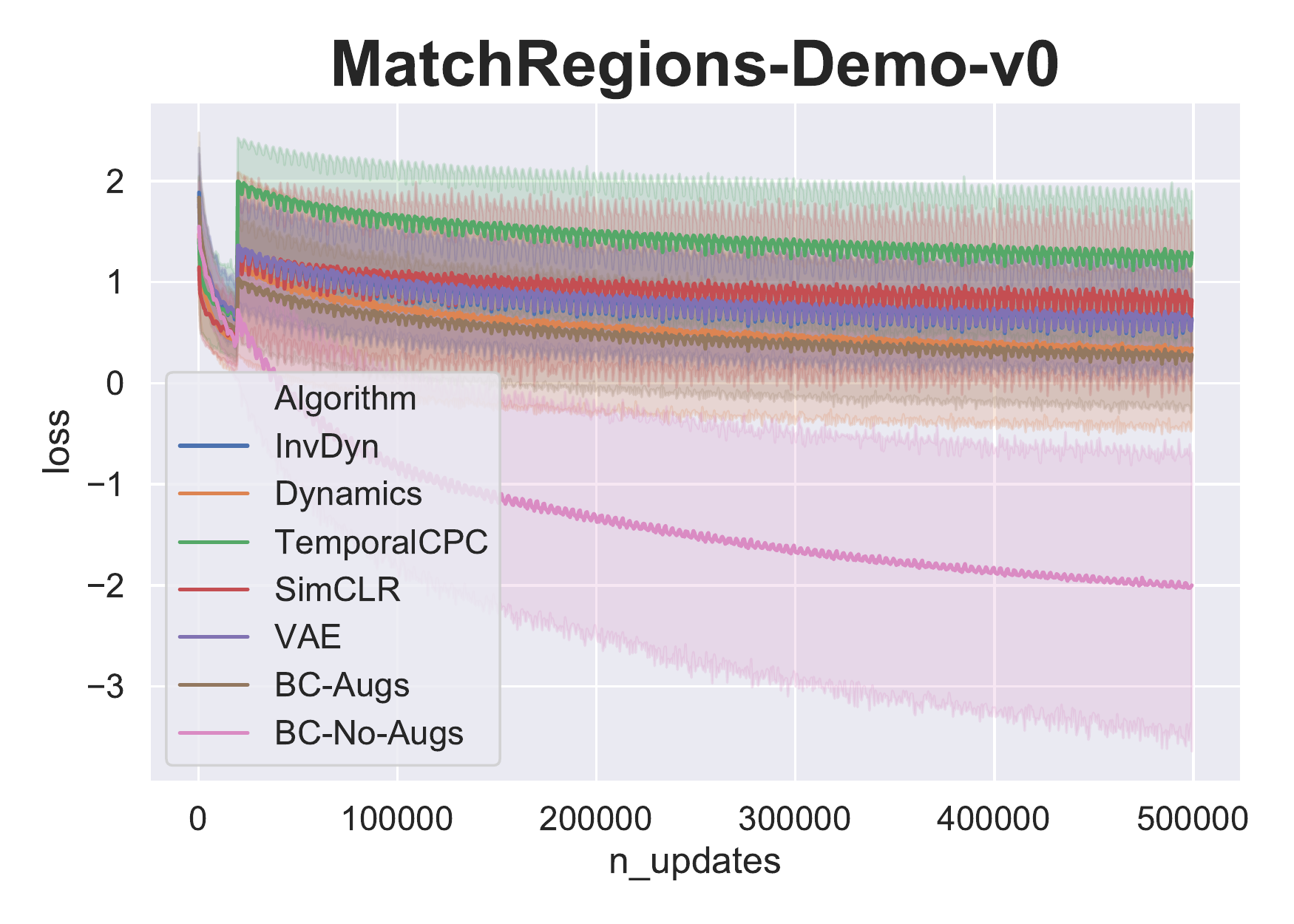}
    \end{subfigure}
    \begin{subfigure}{0.33\textwidth}
        \centering
        \includegraphics[width=\textwidth]{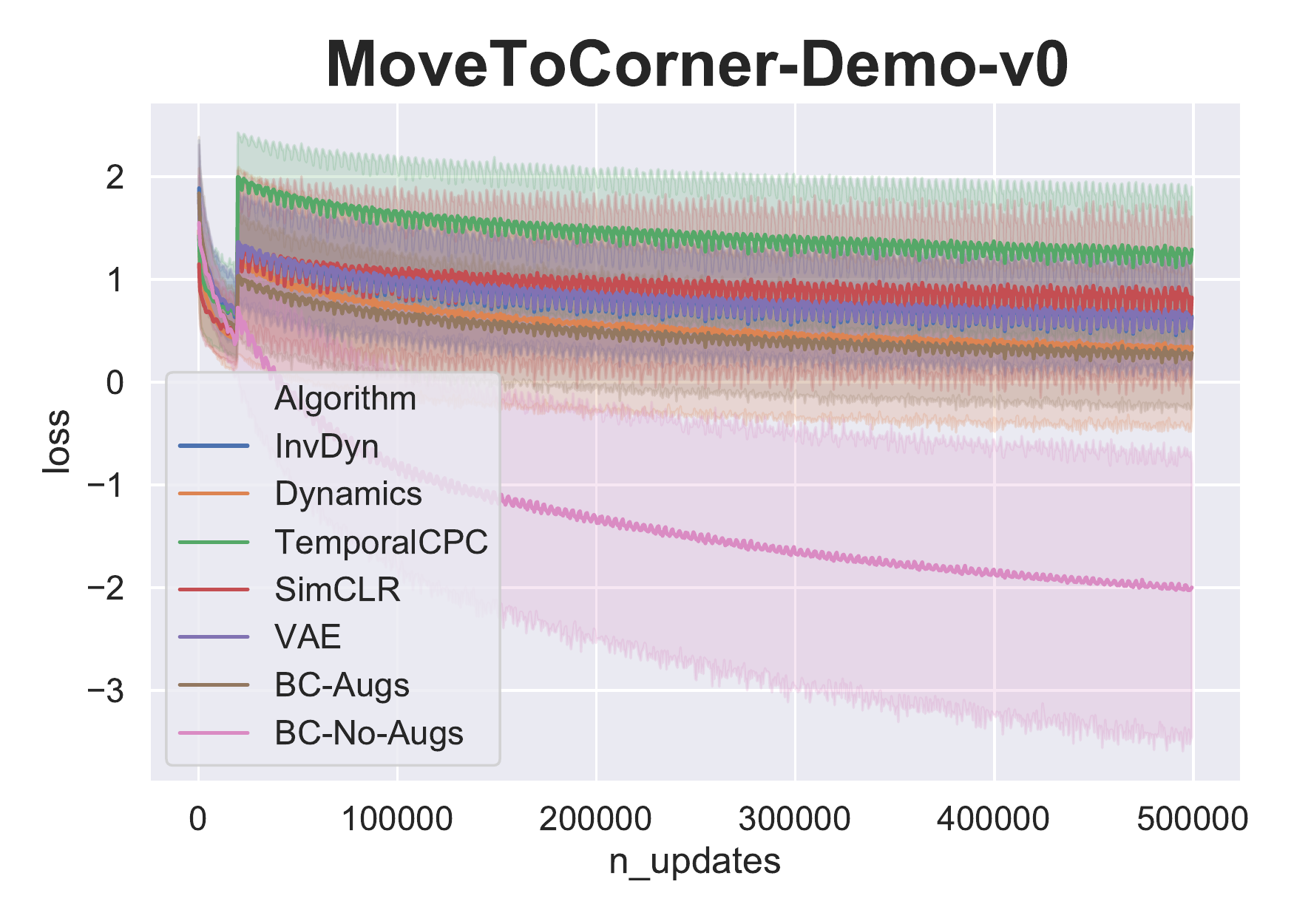}
    \end{subfigure}
    \begin{subfigure}{0.33\textwidth}
        \centering
        \includegraphics[width=\textwidth]{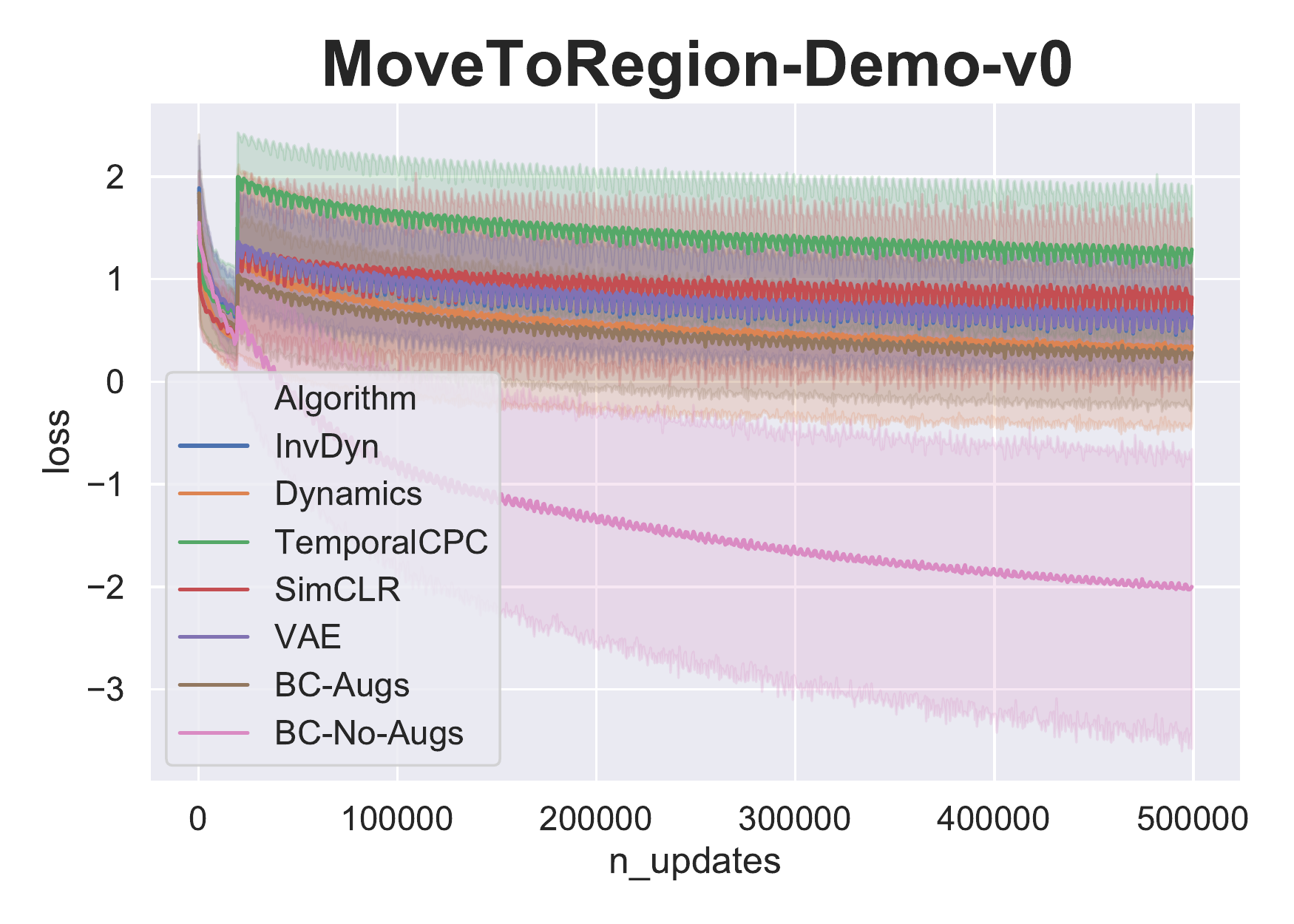}
    \end{subfigure}
\end{figure}
\begin{figure}[H]
    \begin{subfigure}{0.33\textwidth}
        \centering
        \includegraphics[width=\textwidth]{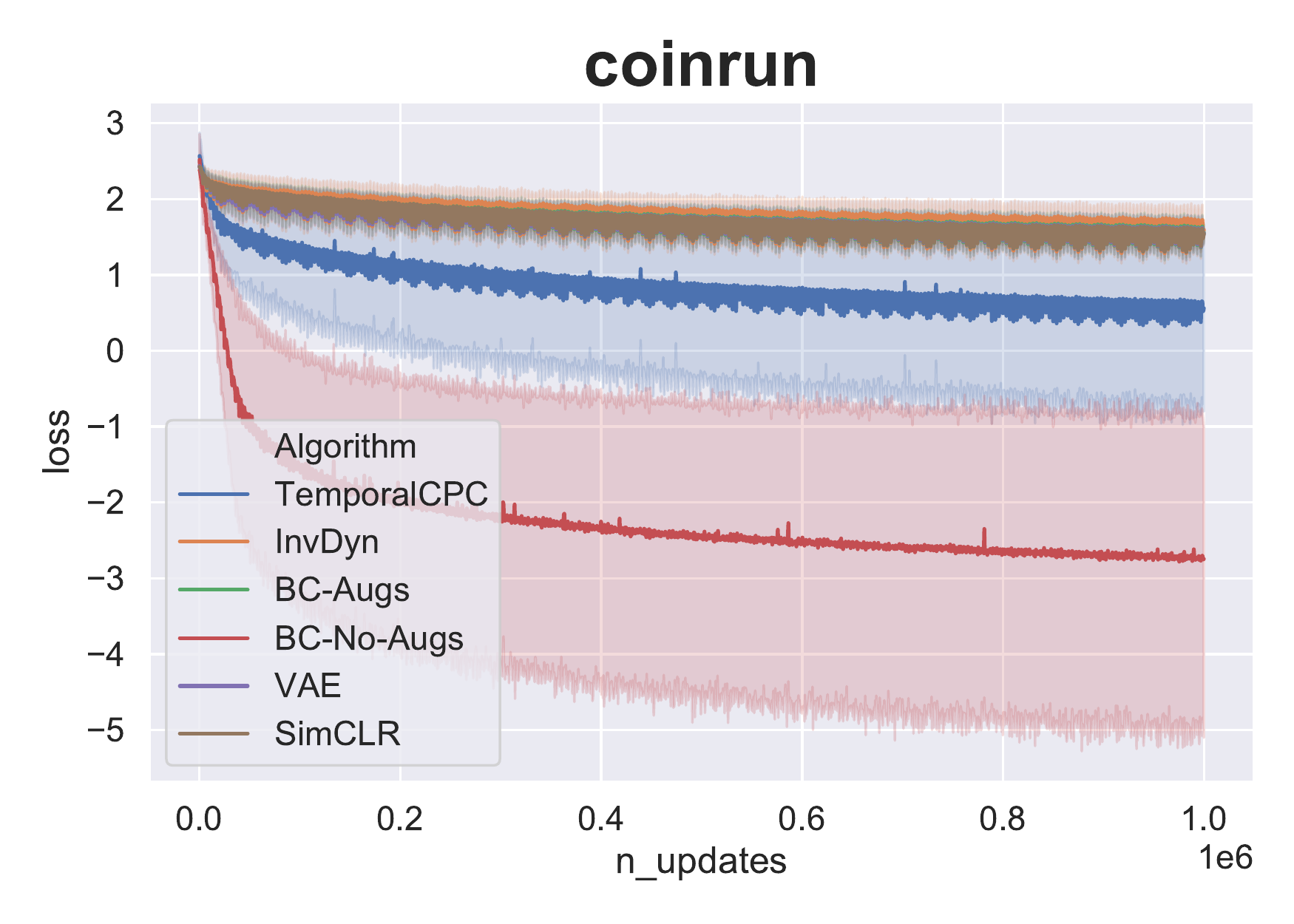}
    \end{subfigure}
    \begin{subfigure}{0.33\textwidth}
        \centering
        \includegraphics[width=\textwidth]{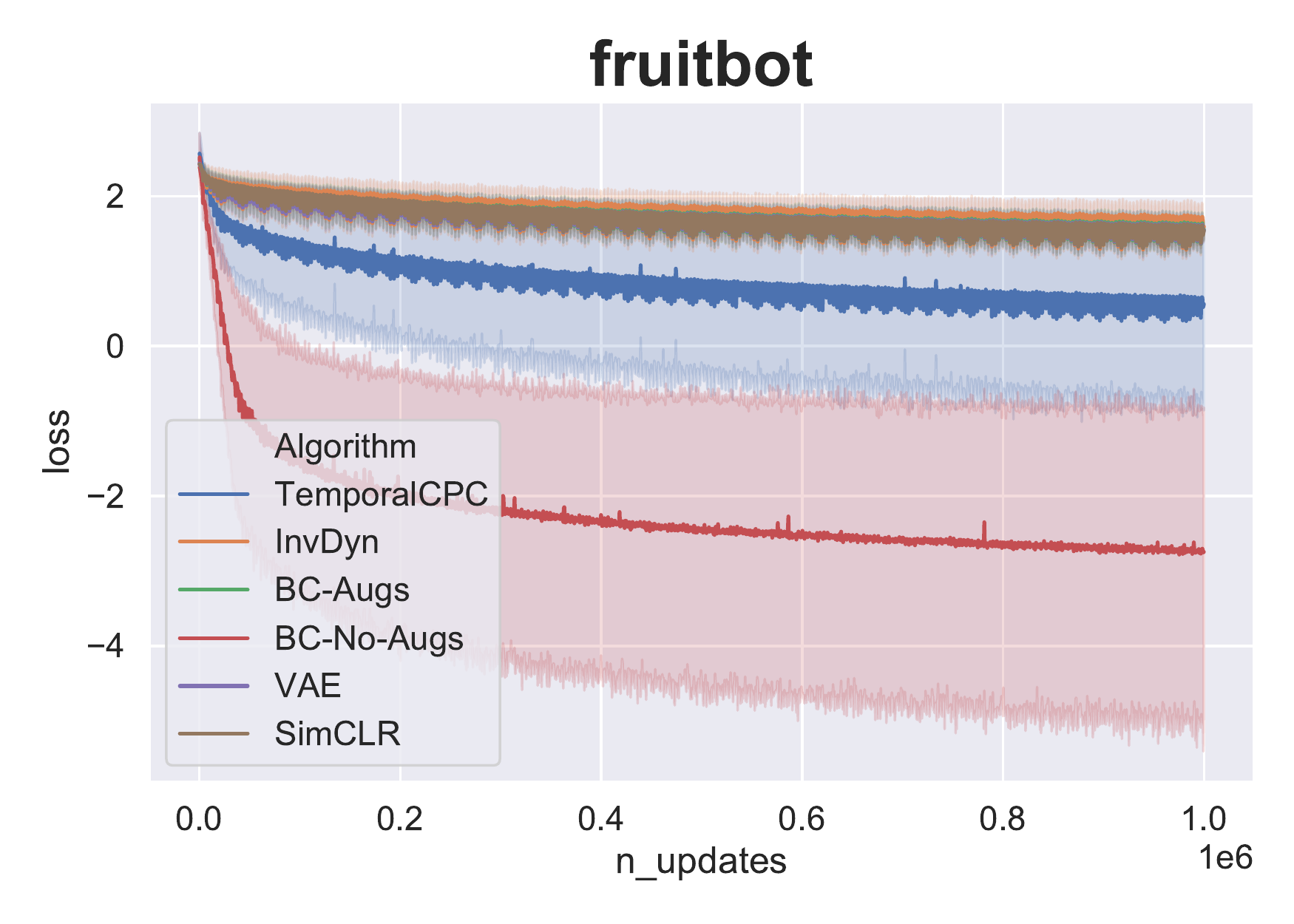}
    \end{subfigure}
    \begin{subfigure}{0.33\textwidth}
        \centering
        \includegraphics[width=\textwidth]{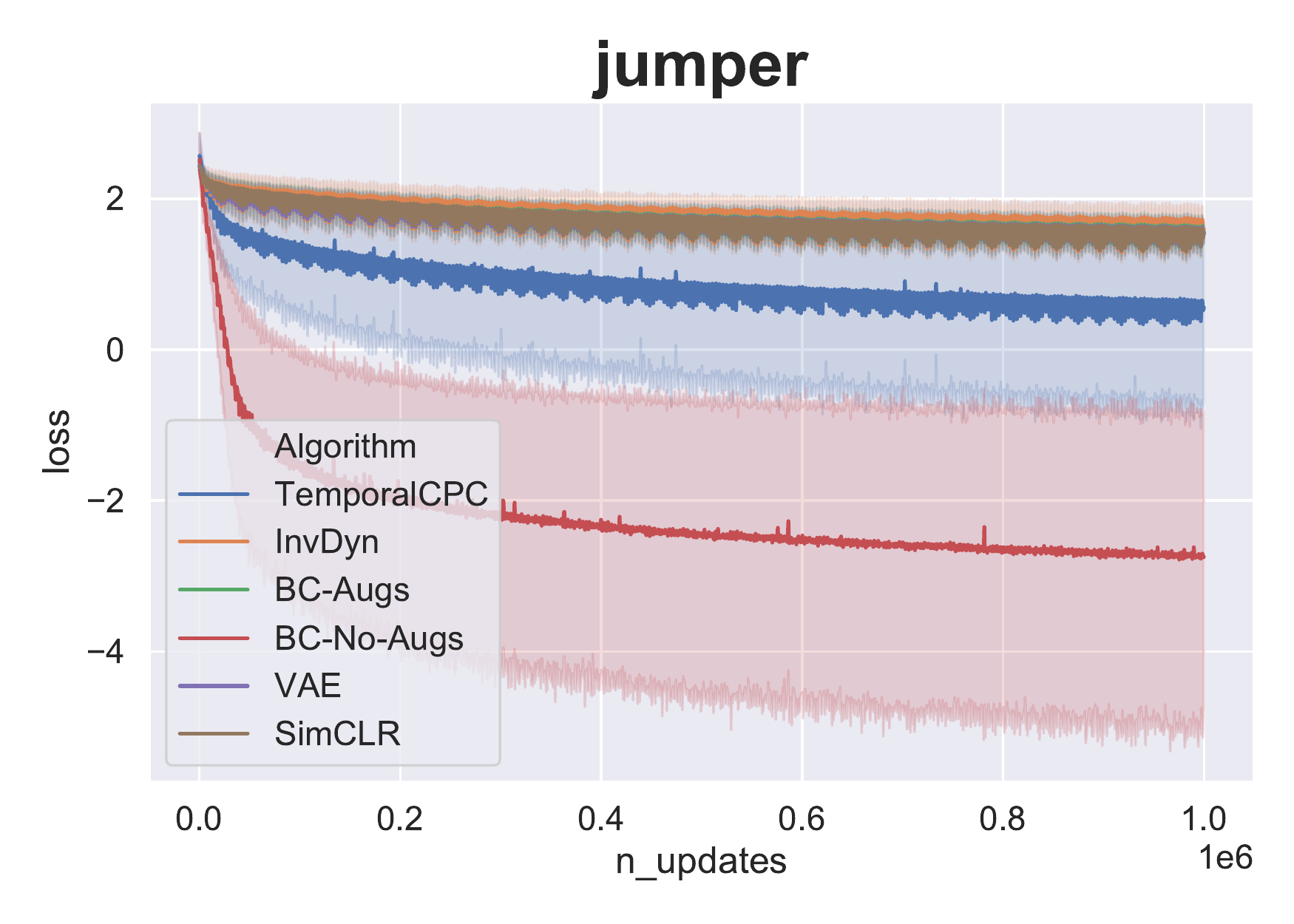}
    \end{subfigure}
\end{figure}
\begin{figure}[H]
    \begin{subfigure}{0.33\textwidth}
        \centering
        \includegraphics[width=\textwidth]{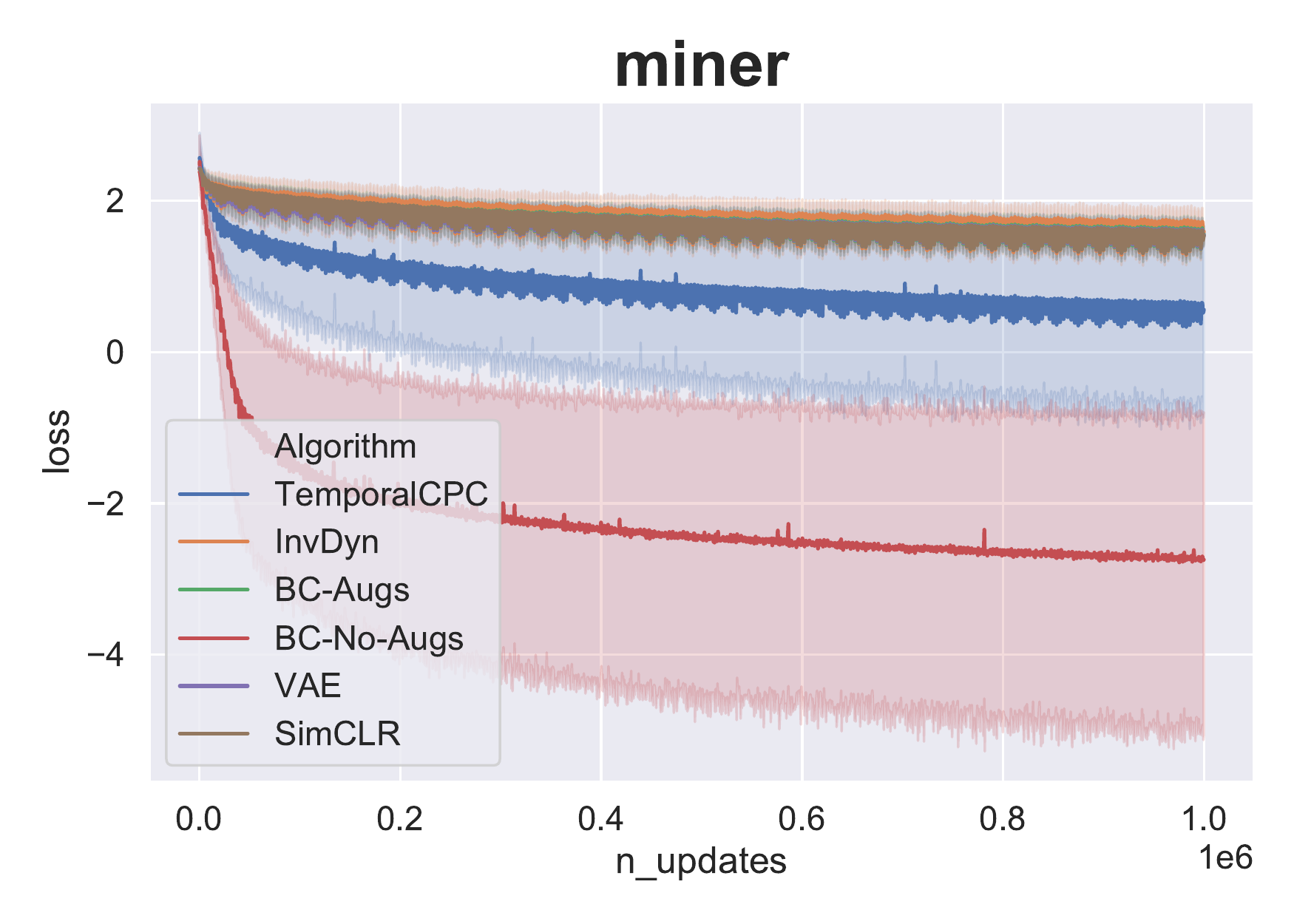}
    \end{subfigure}
\end{figure}

\subsection{Joint Training}
\begin{figure}[h!]
    \begin{subfigure}{0.33\textwidth}
        \centering
        \includegraphics[width=\textwidth]{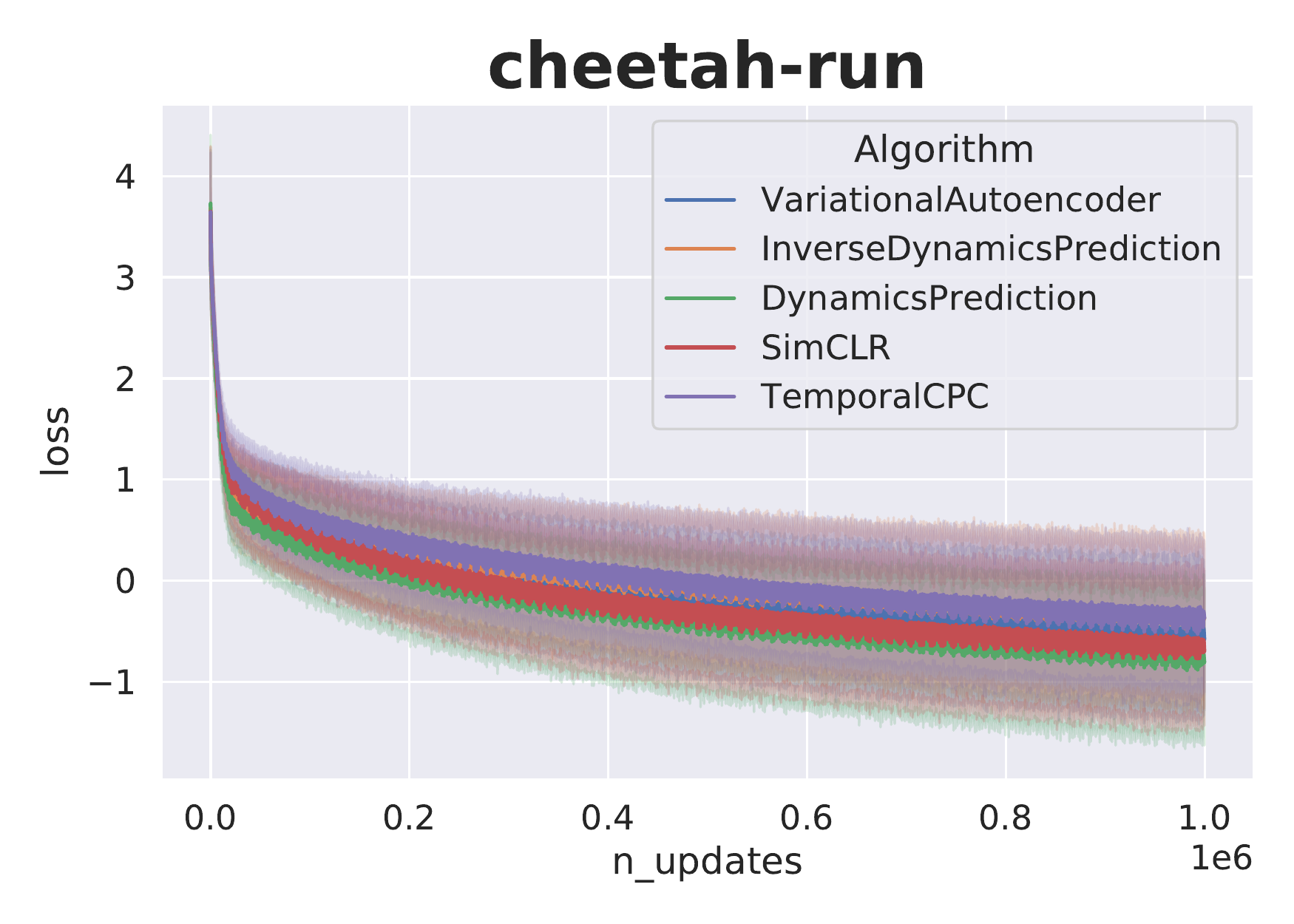}
    \end{subfigure}
    \begin{subfigure}{0.33\textwidth}
        \centering
        \includegraphics[width=\textwidth]{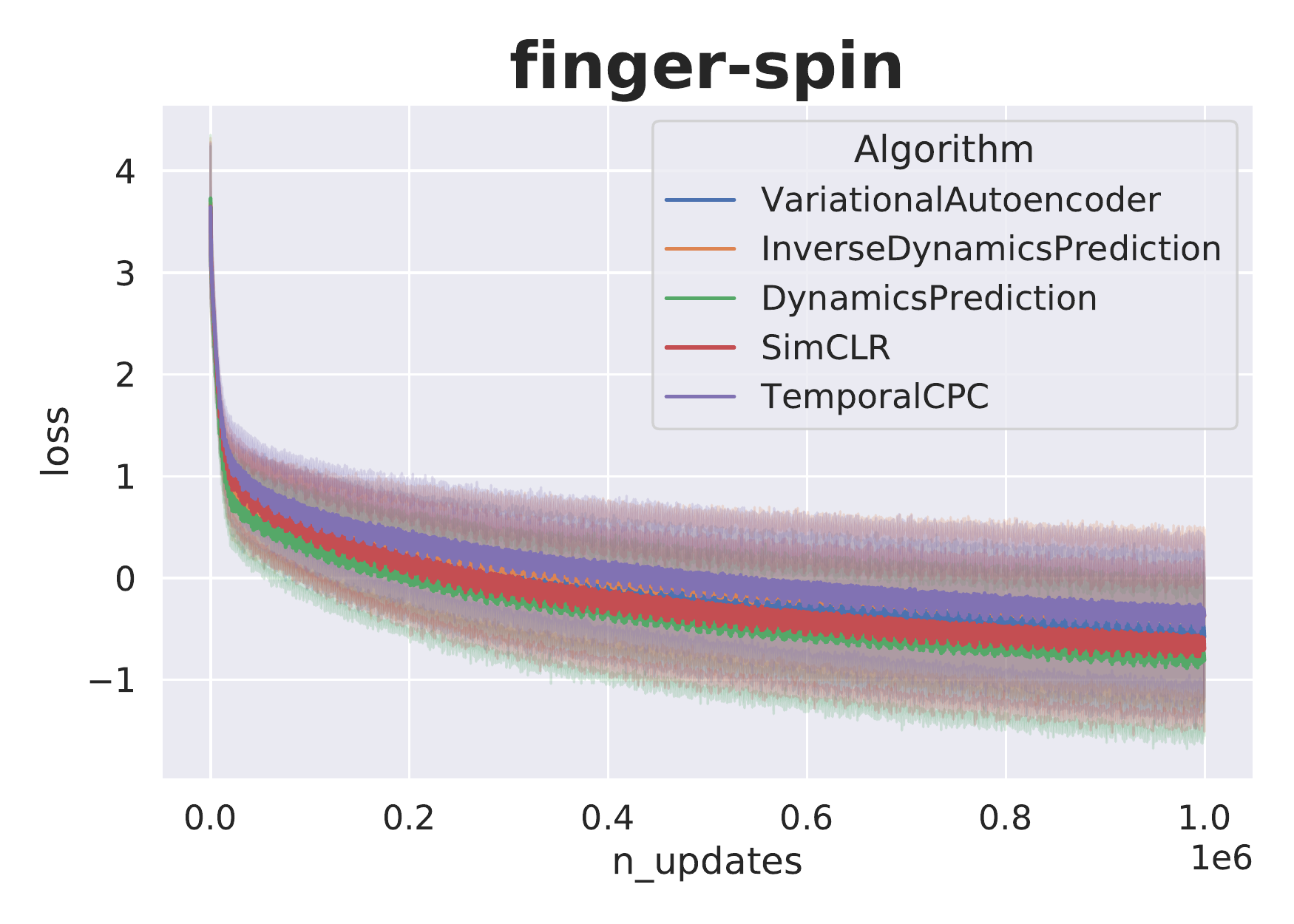}
    \end{subfigure}
    \begin{subfigure}{0.33\textwidth}
        \centering
        \includegraphics[width=\textwidth]{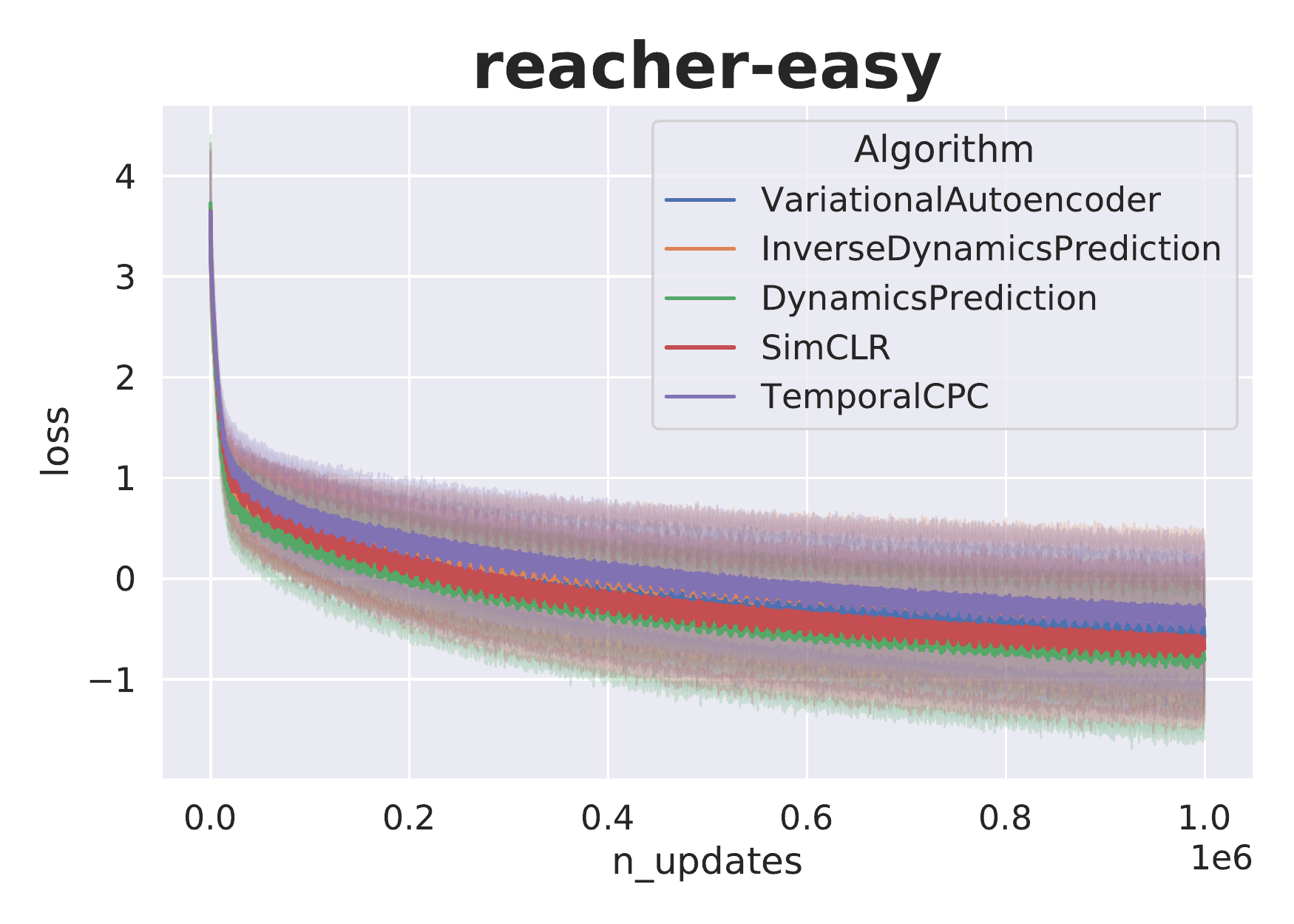}
    \end{subfigure}
\end{figure}
\begin{figure}[h!]
    \begin{subfigure}{0.33\textwidth}
        \centering
        \includegraphics[width=\textwidth]{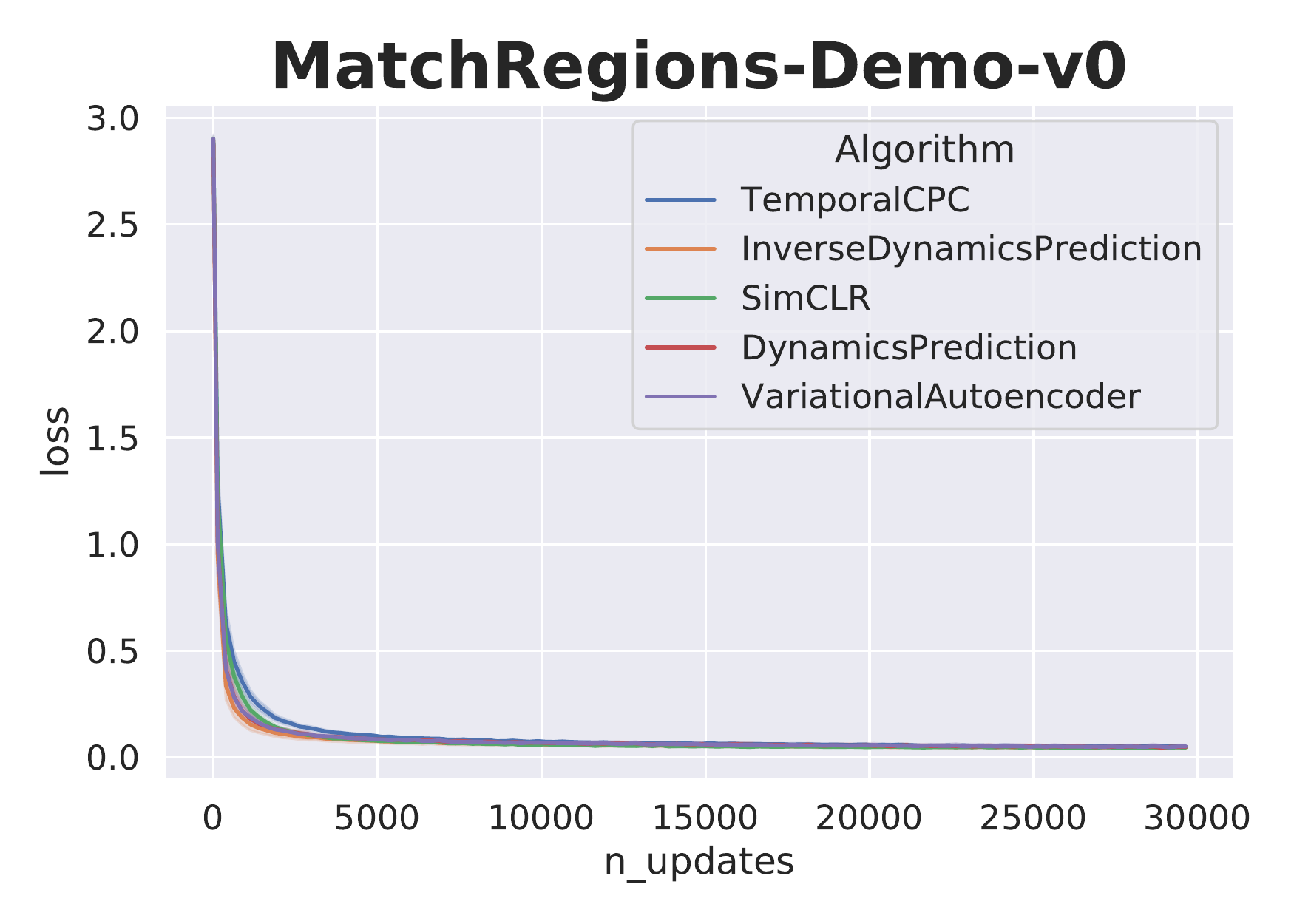}
    \end{subfigure}
    \begin{subfigure}{0.33\textwidth}
        \centering
        \includegraphics[width=\textwidth]{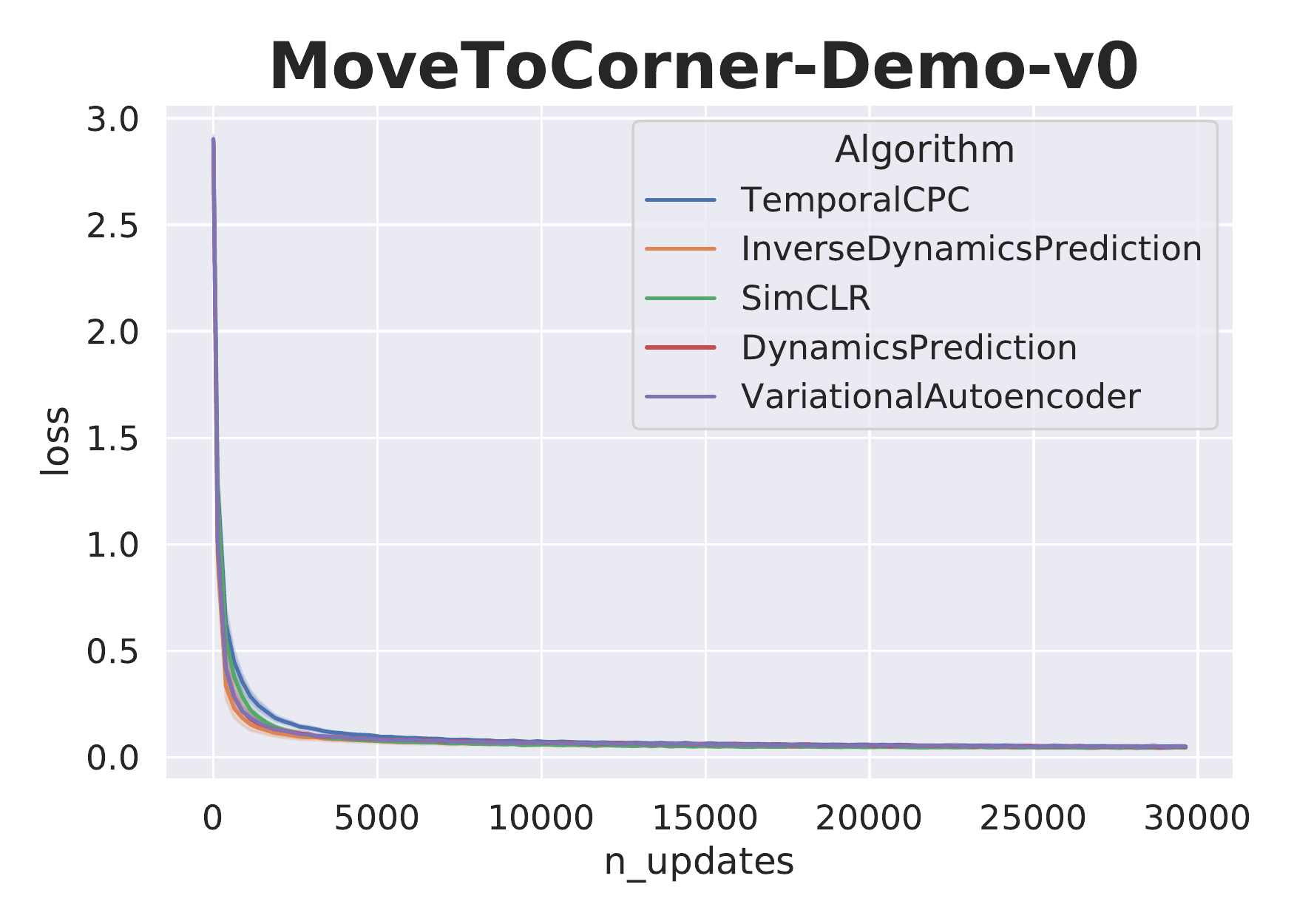}
    \end{subfigure}
    \begin{subfigure}{0.33\textwidth}
        \centering
        \includegraphics[width=\textwidth]{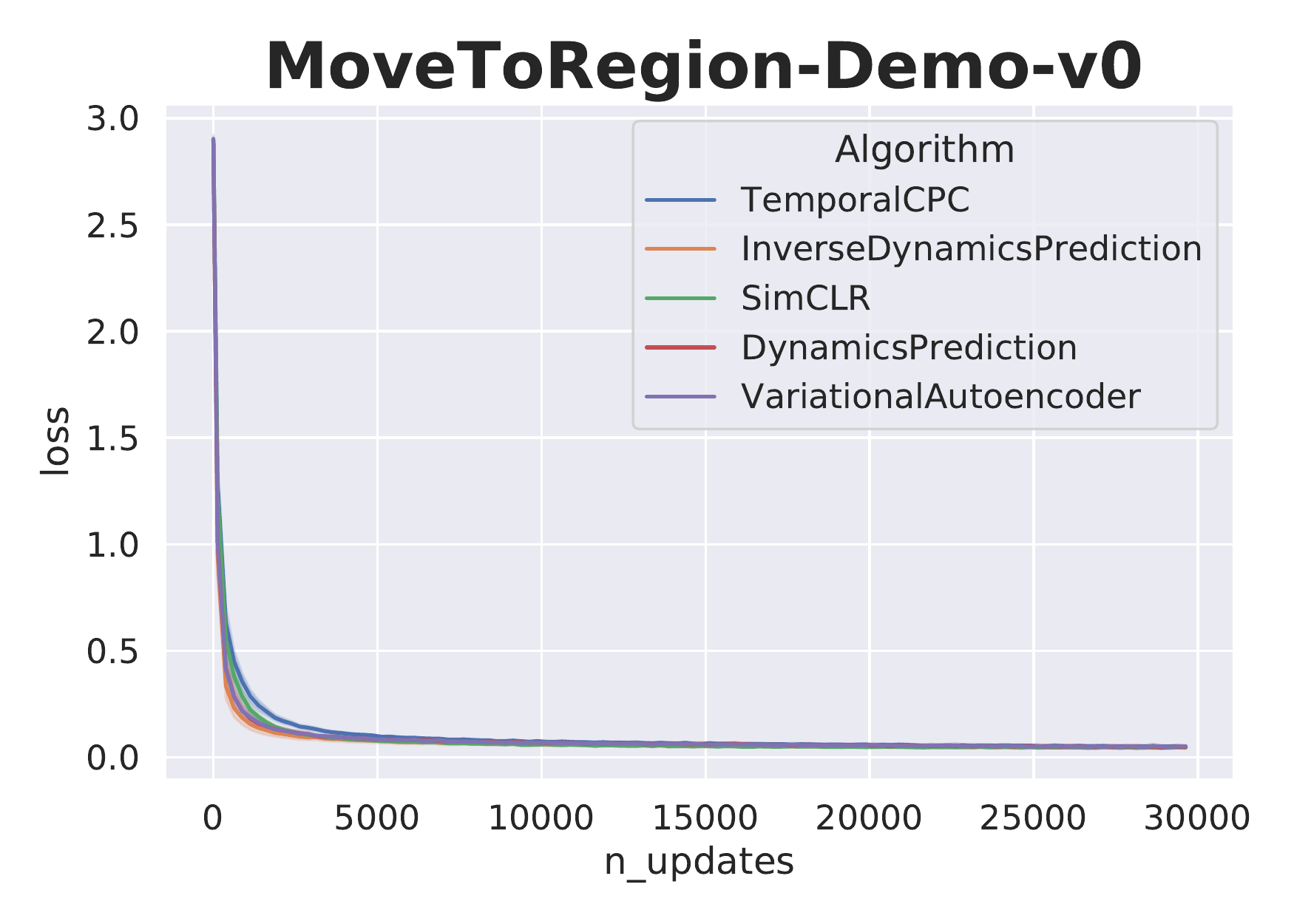}
    \end{subfigure}
\end{figure}
\begin{figure}[h!]
    \begin{subfigure}{0.33\textwidth}
        \centering
        \includegraphics[width=\textwidth]{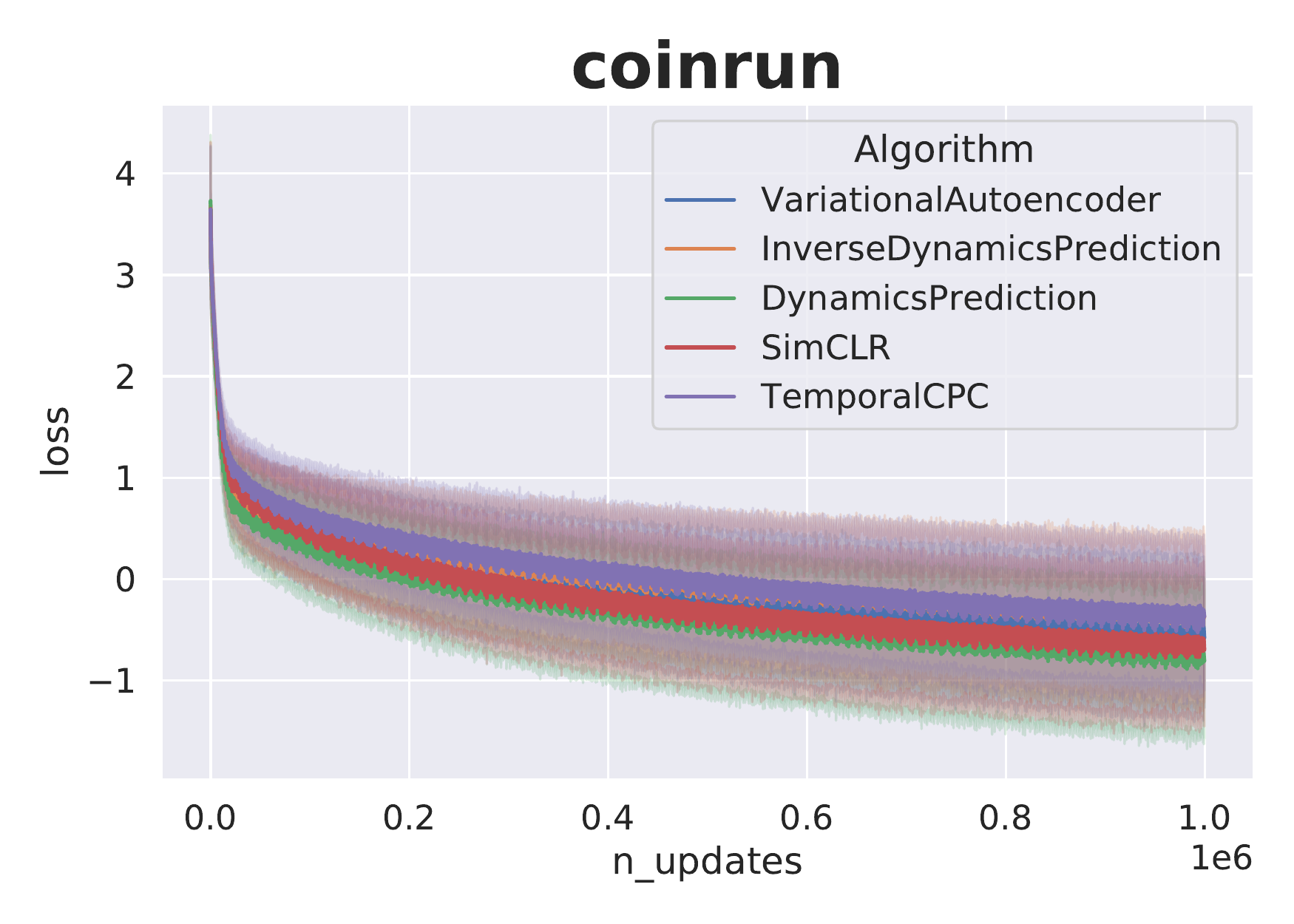}
    \end{subfigure}
    \begin{subfigure}{0.33\textwidth}
        \centering
        \includegraphics[width=\textwidth]{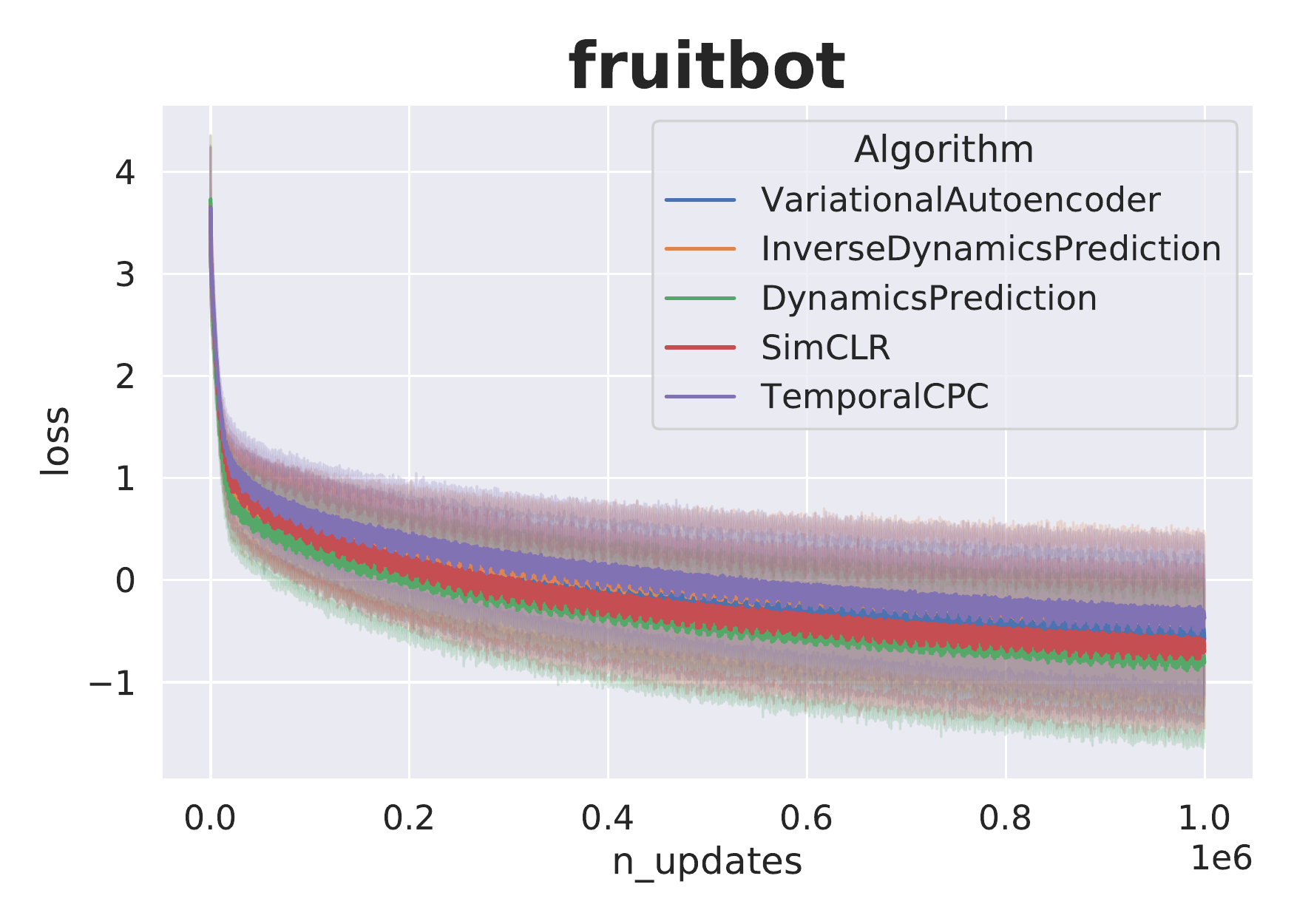}
    \end{subfigure}
    \begin{subfigure}{0.33\textwidth}
        \centering
        \includegraphics[width=\textwidth]{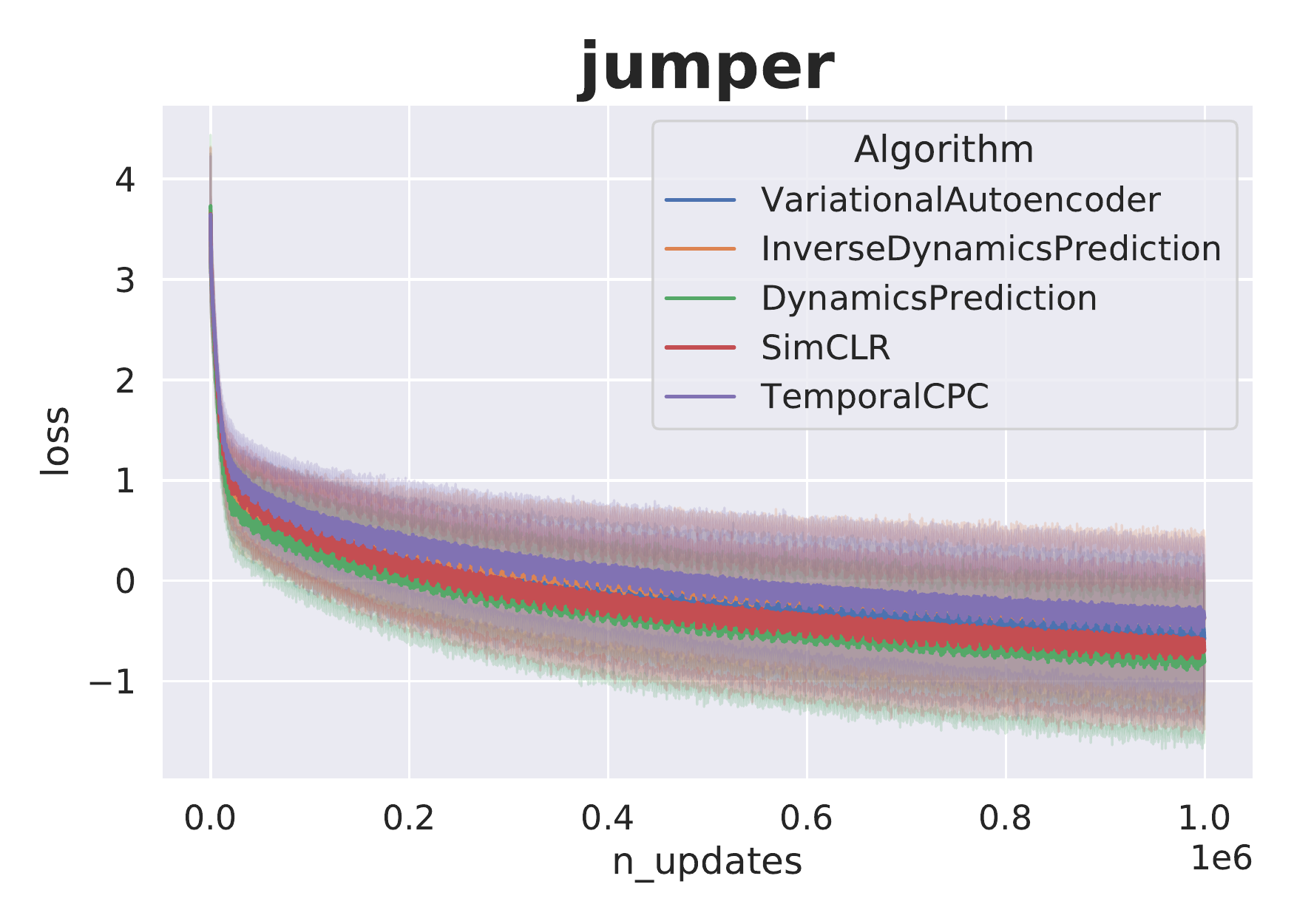}
    \end{subfigure}
\end{figure}
\begin{figure}[h!]
    \begin{subfigure}{0.33\textwidth}
        \centering
        \includegraphics[width=\textwidth]{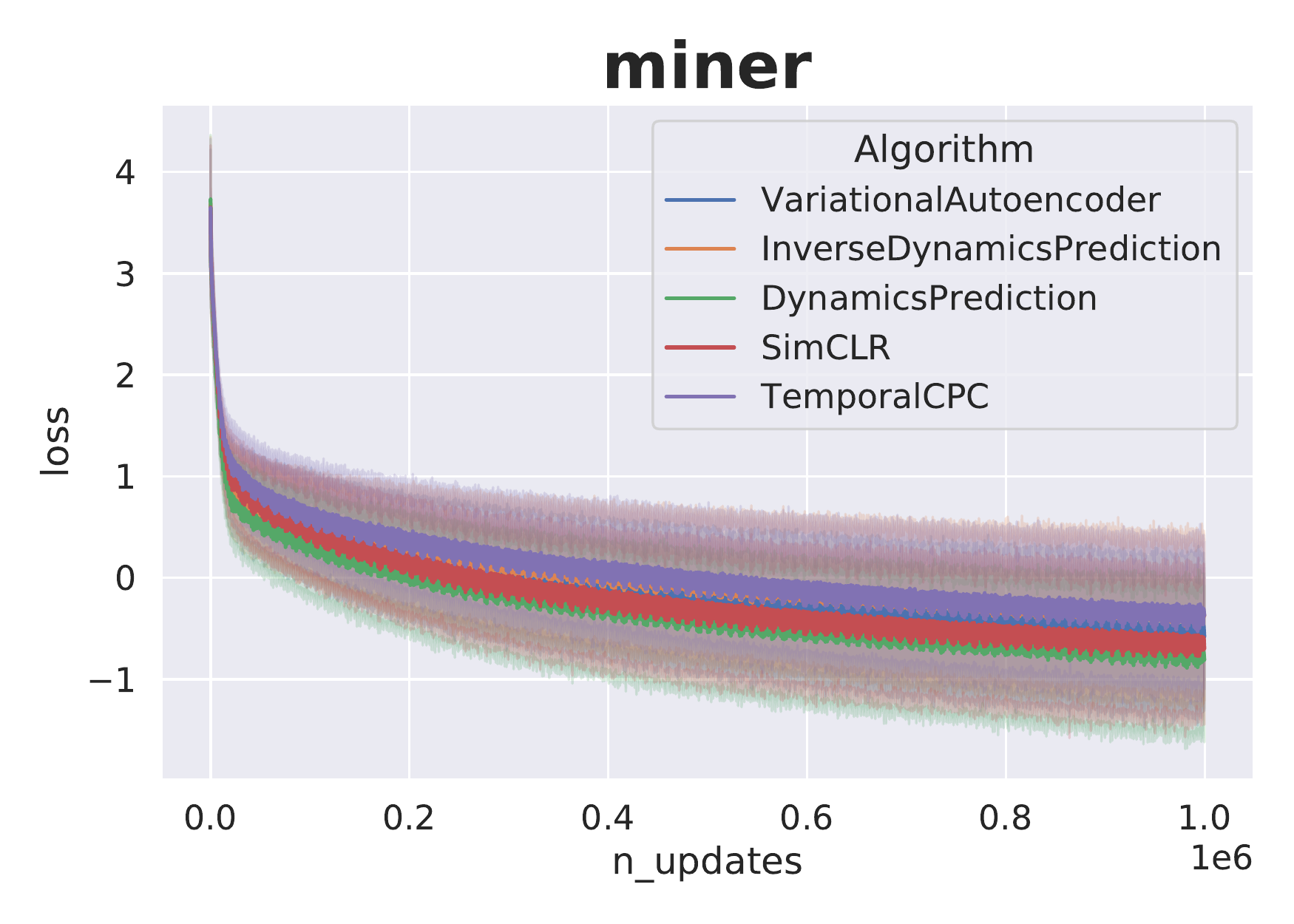}
    \end{subfigure}
\end{figure}

\section{Additional contrastive learning ablations}

\cref{tab:magical-simclr-ablations} presents ablations for SimCLR on our three MAGICAL tasks, using BC as the downstream IL algorithm.
In particular, we experiment with:
\begin{description}
\item[Projection heads] By default, SimCLR uses a ``symmetric'' projection strategy that applies the same projection head to the encoded contexts and encoded target before computing the loss.
We also try using asymmetric projection heads, which are allowed to apply different transforms to the target and context embeddings, and additionally experiment with removing projection heads entirely so that we are computing SimCLR loss directly on the encoder representation.
\item[Compression] We experiment with compression by replacing the default SimCLR loss with the CEB loss~\cite{fischer2020ceb}, but leaving all other training and architecture details the same.
\item[Momentum] In the momentum ablation, we replace the SimCLR loss and encoder with a MoCo-style~\cite{he2020momentum} loss and momentum encoder. Again, all other training and architecture details are left the same as in our SimCLR implementation.
\end{description}
In \cref{tab:magical-simclr-ablations}, we see that none of these modifications significantly improve performance over standard SimCLR.
For this reason, we expect that using different projection heads, using compression, and using momentum are unlikely to affect the conclusions of our work.


\begin{table}[H]
    \centering
    \begin{footnotesize}
        \begin{adjustbox}{center}
            \begin{tabular}{@{}lcccc|c@{}}
                \toprule
                \textbf{Task} & \textbf{Asymm. proj.} & \textbf{No proj.} & \textbf{CEB loss} & \textbf{Momentum} & \textbf{SimCLR} \\
                \midrule
MatchRegions-Demo & \cellcolor[HTML]{fff7df}0.43$\mpm$0.03 & \cellcolor[HTML]{fff7df}0.44$\mpm$0.02 & \cellcolor[HTML]{fff7df}0.42$\mpm$0.03 & \cellcolor[HTML]{fff7df}0.45$\mpm$0.04 & 0.42$\mpm$0.04 \\ 
MoveToCorner-Demo & 0.78$\mpm$0.03 & 0.83$\mpm$0.03 & 0.80$\mpm$0.03 & 0.83$\mpm$0.03 & 0.86$\mpm$0.06 \\ 
MoveToRegion-Demo & \cellcolor[HTML]{fff7df}0.82$\mpm$0.01 & 0.80$\mpm$0.03 & \cellcolor[HTML]{fff7df}0.83$\mpm$0.02 & 0.81$\mpm$0.01 & 0.81$\mpm$0.02 \\ 
                \bottomrule
            \end{tabular}
        \end{adjustbox}
    \end{footnotesize}
    \caption{
      Ablations for SimCLR variants on MAGICAL.
      We used SimCLR as a pretraining step for BC.
      Significance levels were evaluated relative to vanilla SimCLR (the rightmost column) using a one-sided Welch's t-test at $p<0.05$, as with our other results.
      None of these results differ significnatly from SimCLR, and so none are starred.
    }
    \label{tab:magical-simclr-ablations}
\end{table}

\end{document}